\documentclass[journal]{IEEEtran}
\IEEEoverridecommandlockouts
\usepackage{cite}
\usepackage{amsmath,amssymb,amsfonts}
\usepackage{algorithm}
\usepackage{algorithmic}
\usepackage{graphicx}
\usepackage{textcomp}
\usepackage{xcolor}
\usepackage{soul}    
\usepackage{svg}
\usepackage{amsmath}
\def\BibTeX{{\rm B\kern-.05em{\sc i\kern-.025em b}\kern-.08em
    T\kern-.1667em\lower.7ex\hbox{E}\kern-.125emX}}
\begin{document}

\IEEEpubid{
\fbox{\parbox{\dimexpr\columnwidth+5\fboxsep+5\fboxrule\relax}{
    This work has been submitted to the IEEE for possible publication. Copyright may be transferred without notice, after which this version may no longer be accessible.
  }}
}

\title{Multi-agent Assessment with QoS Enhancement for HD Map Updates in a Vehicular Network
\thanks{Manuscript received XXX; revised XXX; accepted xxx. This work was supported by XXXXXXXXXX. 
(\textit{Corresponding authors: Jeffrey Redondo)}

J. Redondo, N. Aslam and J. Zhang are with the Department of Computer and Information Sciences, Northumbria University, NE1 8ST Newcastle upon Tyne, U.K. (e-mail: jeffrey.redondo@northumbria.ac.uk; nauman.aslam@northumbria.ac.uk; juan.zhang@northumbria.ac.uk).

Z. Yuan is with the School of Engineering, University of Warwick, Coventry, CV4 7AL, U.K. (e-mail: Zhenhui.Yuan@warwick.ac.uk).

}}

\author{Jeffrey Redondo,~\IEEEmembership{Student Member, IEEE},  
Nauman Aslam, \IEEEmembership{Member, IEEE,} \\
Juan Zhang, \IEEEmembership{Member, IEEE,}
and 
Zhenhui Yuan, \IEEEmembership{Member, IEEE}
}

\maketitle

\begin{abstract}
Reinforcement Learning (RL) algorithms have been used to address the challenging problems in the offloading process of vehicular ad hoc networks (VANET). More recently, they have been utilized to improve the dissemination of high-definition (HD) Maps. Nevertheless, implementing solutions such as deep Q-learning (DQN) and Actor-critic at the autonomous vehicle (AV) may lead to an increase in the computational load, causing a heavy burden on the computational devices and higher costs. Moreover, their implementation might raise compatibility issues between technologies due to the required modifications to the standards. Therefore, in this paper, we assess the scalability of an application utilizing a Q-learning single-agent solution in a distributed multi-agent environment. This application improves the network performance by taking advantage of a smaller state, and action space whilst using a multi-agent approach. The proposed solution is extensively evaluated with different test cases involving reward function considering individual or overall network performance, number of agents, and centralized and distributed learning comparison. The experimental results demonstrate that the time latencies of our proposed solution conducted in voice, video, HD Map, and best-effort cases have significant improvements, with $40.4\%$, $36\%$, $43\%$, and $12\%$ respectively, compared to the performances with the single-agent approach.
\end{abstract}

\begin{IEEEkeywords}
High definition map, contention window, latency, prioritization, LiDAR, access category
\end{IEEEkeywords}

\section{INTRODUCTION}
\IEEEPARstart{O}NE of the primary challenges in autonomous vehicles (AVs) is realising precise self-localization within their environment to ensure passengers' safety during transit. To achieve this objective, a high-definition (HD) Map provides extra information including around buildings, road markings, traffic signals, dynamic road signals, etc. \cite{hdmap_review_creation}, with centimeter-level accuracy. This accuracy is essential for ensuring safe and high-quality driving experience. HD Map has become a crucial application \cite{nvidea_hdmap} in achieving the higher levels of automation stated by the Society of Automotive Engineers (SAE) \cite{sae_level5} for AVs. However, deployment of HD Map is complicated due to the dynamic nature of road environments, where the constant changes drive the necessary updates to the HD Map, making the HD Maps generation particularly challenging due to the processing requirement of extensive data from sensors such as LiDAR or cameras.
Some researchers have demonstrated that the processing time of tasks can be reduced by offloading the raw data to cloud/fog/edge servers \cite{chameleon, edgeMap, hdmap_processing_time}. It is worth noting that the study conducted in \cite{hdmap_processing_time} reported a remarkable $66\%$ decrease in processing time. In addition, other authors \cite{edgeMap} incorporated and demonstrated how the use of crowd-sourcing, and RL approach in edge computing reduces the processing time for HD Map. 
\IEEEpubidadjcol
To support the offloading of HD Map data, wireless communication systems must be capable of handling heavy traffic and low-latency applications. However, vehicular ad hoc networks (VANET) suffer from the challenge of guaranteeing quality of service (QoS) requirements in dynamic traffic flow environments \cite{performance_analysis_HDMAP}, specifically caused by the increase of packet collisions due to the fixed contention window (CW) \cite{ieee80211_cw}, \cite{ieee80211_cw2}. 
In order to address the challenge of allocating the appropriate CW in a dynamic network, numerous experts have suggested leveraging artificial intelligence (AI) solutions. These studies have investigated a range of methods, such as single-agent \cite{q_learning_fairness, q_learning_edca_policy_RL} and multi-agent RL approaches \cite{knowledge_driven, deep_rl_resource_v2v}. However, recent research has shown a strong preference for multi-agents in wireless network problems due to their superior effectiveness in navigating uncertain and non-stationary environments \cite{multi_agent_survey}.  In these multi-agent setups, individual vehicles can function as agents engaged in various tasks such as data transmission for applications including voice, video, HD mapping, and autonomous driving scenarios. For instance, in \cite{deep_rl_resource_v2v}, each vehicle is designated as an agent, showcasing the effectiveness of a multi-agent approach in enhancing resource allocation within a vehicle-to-vehicle (V2V) communication setting. 
However, previous investigations have either acknowledged the use of the HD Map application \cite{edgeMap} or disregarded it \cite{chameleon, adaptive_cw, adaptive_cw_framework, deep_rl_resource_v2v}. Nonetheless, neither of these solutions has evaluated or considered the effects of incorporating various types of services simultaneously. Additionally, these solutions are limited in scalability when it comes to accommodating new applications, as they are MAC layer-based \cite{performance_analysis_HDMAP, adaptive_edca, q_learning_fairness, cw_deep_RL_ieee802.11ax}, which necessitates standard modifications. To address this, a novel single agent RL QoS coverage-awareness algorithm both non-intrusive and tunable has been proposed in \cite{our_sojourn_single_agent} demonstrating an enhancement in terms of latency and throughput. The solution operates on the application level, thus, it does not require any modification to the existing standard. 
\\
\\
Moreover, as the VANET environment becomes more complex with a higher number of vehicles. The dimensionality of state and action spaces for the single agent also increases as more AVs share information with the agent. In a VANET with IEEE802.11p, the exchange of information might lead to congestion and overwhelm the network's capacity. This is because the available bandwidth might not be sufficient to handle the increased load caused by the exchange of control data (agent's actions, rewards) alongside regular data traffic.  Therefore, it is necessary to expand the work in \cite{our_sojourn_single_agent} into a multi-agent system to reduce the complexity of the single agent in a highly dense and high mobility environment such as VANET with the aim to optimize network performance.  
\subsection{Challenges}
Single-agent RL is widely used to demonstrate the improvement in network performance \cite{q_learning_fairness, q_learning_edca_policy_RL}. Nevertheless, in a highly dynamic environment of large scale, a single-agent approach might suffer from high-dimensional state-action spaces along with the increase in the computational load \cite{multi_agent_resource_management_5G_Wifi6}. To mitigate this limitation, the multi-agent approach is considered to reduce the complexity of the problem and improve the learning process by understanding its local environment and subdividing tasks \cite{multi_agent_resource_management_5G_Wifi6}. Implementing RL as a single or multi-agent approach presents many challenges including high dimensionality in terms of the number of features for the state and action space, and scalability in terms of the number of agents.

\paragraph{High Dimensionality}
High dimensionality is a significant challenge in solving problems with a single-agent approach when agents must control a system with a large state and action space. To overcome this limitation, multi-agent systems are often employed because of their ability to interact effectively with the local environment. For instance, the authors in \cite{knowledge_driven} have proposed a solution where the state space has a high dimensionality. The state space includes task profile, edge computing status, and vehicle speed. This state space will be extensive in the setup of a single agent, thus using a multi-agent in the local environment can reduce the state space size. Moreover, their algorithm is based on an asynchronous advantage actor-critic (A3C), global actor, and critic learning approaches which need to transfer data from vehicles to the global actor-critic. This increases the bandwidth usage in a wireless network. In addition to disseminating an agent's learning experiences across a global network, agents may also need to exchange additional types of information, such as their states, actions, or reward functions. For example, in \cite{deep_rl_resource_v2v}, a distributed approach is implemented where agents share information that is added to the observation of other agents, which increments the state space. Sharing information among agents can increase complexity and network congestion, which can be addressed through reward modelling as suggested in \cite{scalable_reward}. In this approach, the reward function provides valuable information without requiring extra feedback from agents (sharing of information). Moreover, having the same reward function offers the advantage of finding the optimal joint action to maximize the rewards of agents \cite{multi_agent_coordination_book}. In our study, we refer to the utility function in \cite{adaptive_cw} as the reward function, and add penalties to improve the stability that properly describes the network status. Through this action, the dimensionality can be effectively reduced because no information is directly shared among agents. The reward function provides sufficient information about the overall network in terms of latency and throughput. Moreover, it reduces the communication between agents, freeing up wireless channels for more efficient transmission of user plane data instead of the control plane. Overall, this reward mechanism enables a streamlined and optimized network operation.

\paragraph{Scalability}
Could the same single-agent strategies be used in the multi-agent scenarios, more specifically could the multi-agent strategy perform better? This problem has been stated in \cite{multi_agent_the_answer}. Thus, similar to the increase in dimensionality, a challenge arises when scaling the solution in multi-agent systems due to the sharing of information between agents (in both state and action spaces). This potentially increases overhead due to the exponentially large number of agents \cite{scalable_multi_agent}. Therefore, to elaborate on the design of an RL multi-agent algorithm for VANET, several questions arise: How many agents must be selected for a VANET scenario? Could each service (Voice (VO), Video (VI), Best-Effort (BE), HD Map) be set as an agent? Or
would having each AV as an agent be more efficient? These are questions addressed in our work.

\paragraph{Compatibility and Computational Capacity}
There have been efforts to address the challenge of CW assignment, yielding positive outcomes in terms of latency, throughput, and fairness. However, translating these findings into practical implementation is more complex than it may initially appear, requiring modifications to the existing standard. Additionally, most solutions work with deep reinforcement learning (DRL) policy gradient, advantage actor-critic (A2C), and A3C, which increases the computational cost and might further overload the Onboard Unit (OBU) of the AV. Thus, our solution operates on the application layer to solve the compatibility problem. Specifically, it utilizes a Q-learning approach to reduce the computation capacity requirements. 

\subsection{Contributions}
The main contributions of this paper are summarized as follows:
\begin{itemize}
    \item A novel distributed and lightweight multi-agent solution that effectively improved QoS in IEEE802.11p networks for HD Map is proposed. The solution employs Q-learning and the same reward function for each agent while addressing the high dimensionality issue and reducing computational complexity. The proposed approach ensures QoS while demonstrating a significant improvement in overall performance.
    \item Two distinct multi-agent setups, which demonstrate the versatility of multi-agent systems in allocating wireless resources to ensure QoS, are evaluated. Additionally, a distributed multi-agent solution that exceeds the performance of centralized single-agent approaches is also introduced.
    \item The performance between centralized and distributed learning in wireless resource allocation is extensively assessed. We delve into the intricacies of these approaches to understand their respective advantages and drawbacks. Thus, we explored the implication of implementing the machine learning solution directly on the AVs or at the edge server.  This analysis offers valuable insights into the current wireless resource allocation optimization area of investigation. It provides an understanding of the benefits associated with RL in VANET to offer QoS. 
\end{itemize}
\section{Related Work}
To enhance the QoS in the standard IEEE802.11, extensive research has been conducted to improve latency by developing new solutions to allocate CW value and the enhanced distributed channel access (EDCA) mechanism. Approaches such as new access categories, single-agent RL, and multi-agent RL  are commonly used in solutions to provide a better QoS, which are discussed in the following subsections.  

\subsection{CW, EDCA, and single agent}
There are different ways to improve the performance of a wireless network using the family standard IEEE802.11. One approach is to enhance the EDCA mechanism or the selection of the CW. To achieve this, various solutions have been proposed. For instance, some authors suggested new access categories \cite{avaq_edca_new_ac} by extending the number of access categories (AC) from four to seven for different video resolution qualities. Other authors \cite{low_latency_new_ac} added a low latency AC. 

In \cite{performance_analysis_HDMAP} a new AC for HD Map is added to the EDCA showing an improvement in high-density vehicular network. Nevertheless, it is not adaptable to changes in the environment such as the number of vehicles. Besides its implementation requires a change in the standard which is not feasible. Other solutions focus on queuing management \cite{dynamic_queue,q_learning_edca_policy_RL,queue_IoV,logical_EDCA}. Nevertheless, these solutions do not consider the tuning of the parameters CW, which plays a crucial role in enhancing the QoS.

Thus, to improve further the QoS, algorithms are developed to select the proper values for the MAC layer parameter of the standard IEEE802.11 by proposing approaches such as Markov model for varying CW \cite{cw_cooperative_commu},  game theory \cite{adaptive_cw}, AI/ML reinforcement learning Q-learning \cite{q_learning_fairness, RL_cw_simple, fairness_bidirectional_qlearning}, policy gradient \cite{q_learning_edca_policy_RL}, deep RL \cite{adaptive_cw}, \cite{cw_deep_RL_ieee802.11ax}. 

Nevertheless, these solutions have one common requirement of modifying the current standard for their implementation, which is not always feasible for worldwide deployment. To overcome this, in \cite{our_sojourn_single_agent}, we have developed a solution that operates at the application level offering similar priorities that the EDCA and demonstrated an improvement of $10\%$ compared with the new HD Map AC \cite{performance_analysis_HDMAP} in term of latency.

\subsection{Multi Agent Systems for HD Map}
One of the challenges to overcome in the generation of HD Map is the constraints of wireless transmission resources in a dynamic network \cite{edgeMap}. To address this, the authors in \cite{edgeMap} proposed a Distributed Adaptive Offloading and Resource reservation (DATE) solution, which uses multi-agent DRL with a global policy network. Results showed an improvement in the overall latency for offloading. Nevertheless, different types of services were not included, and throughput was also not considered as part of the optimization problem. Another study \cite{knowledge_driven} utilized also a multi-agent approach RL-based scheme in a centralized online training manner with an Asynchronous Advantage Actor-Critic (A3C) algorithm. Both studies have implemented a multi-agent with a disadvantage in terms of computational cost, related to the use of a more complex RL algorithm that might require an increase of the computational cost and overhead because of the deep neural networks and the parameters sharing from each vehicle to the global network, respectively.

To avoid the overhead of a centralized approach, authors in \cite{deep_rl_resource_v2v} proposed a decentralized multi-agent solution to manage the wireless resource allocation, which inspires us to implement and test our single-agent solution \cite{our_sojourn_single_agent} in a distributed multi-agent manner. One key difference between our approach and the mentioned solution is that their reward function includes the sum of the current capacity (Shannon-Hartley Theorem) of each vehicle \cite{deep_rl_resource_v2v}, which indicates that there is an intercommunication between agents so there would be congestion on the network.

In summary, previous solutions fail to consider the integration of different service types, such as voice, video or any other service simultaneously, which are frequently used by end-users. Therefore, it is necessary to expand the computational capacity of AVs by using DRL. We first formulated multiple services, throughput and latency into our optimization problem. Then, we implemented a Q-learning algorithm due to its advantages of requiring less computational load. Thirdly, we utilized the same reward function that includes latency and throughput to avoid information sharing between agents or the global network.

\section{Problem statement}
The VANET environment can be classified as an unknown and partially observable Markov decision process (MDP). This can be represented as a tuple $(V, S, A, R, T)$, where $V$ is the set of AVs, $S$ is the set of states, $A$ is the set of actions, R represents the reward, and $T$ represents the time.  

The environment comprises a dynamic vehicular traffic flow with AVs denoted by set $\mathcal{V} = \{1,...,N\}$. Each vehicle follows a defined route entering and exiting the environment, mimicking the typical dynamics of an urban area where AVs start a route and finish at the destination. Thus, the vehicles do not stay in the environment for the whole simulation. The vehicles are categorized based on their data transmission type (e.g. voice, video, HD map, and best-effort) described by the set of categories $\mathcal{C} = \{1,...,M\}$.

To ensure specific network performance for each service category, the ultimate goal of our work is to provide and maintain the required throughput and latency for all services. Therefore, we consider throughput $\mathcal{R}$ and latency $\mathcal{L}$ in the utility function \cite{adaptive_edca}. Besides, we also incorporate penalties and bonuses, defined as $\mathcal{F}$, into our solution to improve stability, as shown in Eq. \eqref{eq:utility_function_penalty}. 
\begin{equation}
    \begin{split}
    U(c) = \alpha_1\frac{\mathcal{R}(c)}{\mathcal{R}_{max}(c)}- \alpha_2\frac{\mathcal{L}(c)}{\mathcal{L}_{max}(c)} + \mathcal{F}
    \label{eq:utility_function_penalty}
    \end{split}
\end{equation}
where the weights $\alpha_1$ and $\alpha_2$ provide a trade-off between $\mathcal{R}$ and $\mathcal{L}$.
Then the maximization problem can be formulated as follows:
\begin{equation}
    \underset{w_{v,t}}{\max}  \sum_{v \in \mathcal{V}} \sum_{t \in \mathcal{T}} x_{v,t} U_{v,t}(c), \quad \forall c \in \mathcal{C}, v \in \mathcal{V}, t \in \mathcal{T}
    \label{eq:max_problem}
\end{equation}
subject to
\begin{equation}
    x_{v,t} \in \{0,1\}
    \label{eq:constraint_x}
\end{equation}
\begin{equation}
    \frac{1}{|\mathcal{V}|} \sum_{v=1}^{|\mathcal{V}|} \mathcal{L}_{v}(c) \leq \mathcal{L}_{\text{max}}(c), \quad  \mathcal{L} \in \mathbb{R}
    \label{eq:constraint_L}
\end{equation}
\begin{equation}
    \sum_{v=1}^{|\mathcal{V}|} \mathcal{R}_{v}(c) \geq \mathcal{R}_{\text{min}}(c), \quad  \mathcal{R} \in \mathbb{R}
    \label{eq:constraint_R}
\end{equation} 
\begin{equation}
    w_{v,t}(c) \leq w_{\text{max}}(c), \quad w \in \mathbb{R}, \quad  w \neq 0
    \label{eq:constraint_w}
\end{equation}
where $(\ref{eq:constraint_L})$ and $(\ref{eq:constraint_R})$ indicate the maximum latency and minimum data rate per service type, respectively. The last constraint, described in $(\ref{eq:constraint_w})$, is the maximum waiting time allowed per category. The variable $x_{v}$ is a binary index which can be either 0 or 1, indicating whether a vehicle is allowed to transmit.

As described in \cite{our_sojourn_single_agent}, the problem is difficult to solve analytically due to the nonlinearity and the direct relationship between the CW value and $\mathcal{L}$, as well as the inverse proportionality between the CW value and $\mathcal{R}$. Therefore, To address these complexities and find an optimal solution, we employ RL which is particularly adept at uncovering intricate patterns and is crucial in the context of this new multi-agent investigation. Here, each agent seeks to maximize its utility, leading to the complexity of the problem. The agents will learn and select the optimal $w_t$ waiting time before the next data transmission. The notations used in this paper are summarized in Table \ref{tab:notation}.

\begin{table}[h]
\caption{Annotation Table}
\centering
\begin{tabular}{c|c|c|c}
\hline
Variable & Definition & Variable & Definition \\
\hline
$\alpha_1$, $\alpha_2$ & Coefficients   & $slope$ & Slope for Direction \\
$\mathcal{A}$ & Action Set              & $\mathcal{T}$ & Time Set \\
$\mathcal{C}$& Category Set             & 
$T_v$ & Veh Active (RSU) \\
$d$& Distance                           & 
$T_{cv}$ & Veh Active per $\mathcal{C}$\\
$\mathcal{F}$ & Penalty and Bonuses     & $U$ & Utility Function\\

$J$ & Fairness Index                    &  $w$&  Waiting Time\\
$\mathcal{L}$ & Latency                 & $w_{max}$& Max Waiting Time \\
$\mathcal{L}_{max}$& Max Latency        & $x$& Binary Index\\
$\pi$& Policy                           & $\mathcal{R}$& Data Rate\\
$S$&State Space                         & $\mathcal{R}_{min}$& Min Data Rate\\
$S_j$& Discrete Sojourn Time             & $\mathcal{V}$ & Vehicle Set\\
$s_j$ & Sojourn Time                     & $\epsilon$&Epsilon-greedy \\
\hline
\end{tabular}
\label{tab:notation}
\end{table}

\section{Proposed Solution Design}

\subsection{Reinforcement learning}
In this comprehensive investigation, the Q-learning approach selected in \cite{our_sojourn_single_agent} has been extended into a multi-agent environment. This approach is based on the Q-Learning Temporal Difference (TD) model-free method offering advantages from dynamic programming and the Monte Carlo method. For instance, the TD method requires less computational resources because it updates estimates incrementally, one step at a time, rather than updating all estimates after every episode. The Q value update is defined by Eq. \eqref{eq_q_learning}.

\begin{equation}
    Q(s,a) = Q(s,a) + \alpha\left[r+\gamma*max_{a'}Q(s',a')-Q(s,a)\right]
\label{eq_q_learning}
\end{equation}

As stated in \cite{multi_agent_coordination_book}, it is advantageous to use the same reward function in finding an equilibrium in a cooperative Multi-agent MDP (MMDP). Therefore, we adopt this approach to improve the learning process. Besides, each agent is categorized as an independent learner (IL) to generate a smaller Q-table compared to the joint state-action space, $S \times A^m$. For simplicity, $y$ is defined as $y=\gamma*max_{a'}Q(s,a)-Q(s,a)$, and by replacing the utility function Eq. \eqref{eq:utility_function_penalty} as the reward $r$ into Eq. \eqref{eq_q_learning} then, the updated equation is described as follows:
\begin{equation}
\begin{split}
        Q(s,a) = Q(s,a) + \alpha[U + y]
    \label{eq_q_learning_2}
\end{split}
\end{equation}

\subsubsection{State}
The set of state $\mathcal{S}$ is defined as follows \cite{our_sojourn_single_agent}:
\begin{equation}
    S = \{S_j, T_v, C, T_{cv}\}
\end{equation}
where $S_j$ is the sojourn time, during which a vehicle is within the coverage area of the Roadside Unit (RSU). In the measurement, 0 means the least time and 4 is the longest time. The sojourn time is converted into five discrete values from 0 to 4, for more detail check \cite{our_sojourn_single_agent}. The value $T_v$ is the total number of vehicles active with a maximum of $N$, it is obtained at the edge server side by the Algorithm 1 in \cite{our_sojourn_single_agent}. The set of categories $C$ in this investigation refers to the four services voice, video, HD map, and best-effort. Finally, $T_{cv}$ stands for the total number of active vehicles per category $c$ counted by Algorithm 3 in \cite{our_sojourn_single_agent}.

\begin{equation}
\Phi = \left\{
\begin{aligned}
    &Q^1 \\
    &Q^2 \\
    &\vdots \\
    &Q^n
\end{aligned}
\right\}
=
\left\{
\begin{aligned}
    & \mathbb{E}^{\pi^1}[R|S=\{S_j, T_v, C, T_{cv}\},a] \\
    & \mathbb{E}^{\pi^2}[R|S=\{S_j, T_v, C, T_{cv}\},a] \\
    &\vdots \\
    & \mathbb{E}^{\pi^n}[R|S=\{S_j, T_v, C, T_{cv}\},a]
\end{aligned}
\right\}
\label{eq:multi_agent_equation}
\end{equation}

\begin{equation}
\Phi(c) = \left\{
\begin{aligned}
    &Q^1 \\
    &Q^2 \\
    &\vdots \\
    &Q^n
\end{aligned}
\right\}
=
\left\{
\begin{aligned}
    & \mathbb{E}^{\pi^1}[R(c)|S=\{S_j, T_v, T_{cv}\},a] \\
    & \mathbb{E}^{\pi^2}[R(c)|S=\{S_j, T_v, T_{cv}\},a] \\
    &\vdots \\
    & \mathbb{E}^{\pi^n}[R(c)|S=\{S_j, T_v, T_{cv}\},a]
\end{aligned}
\right\}
\label{eq:multi_agent_equation2}
\end{equation}

Observing the state can reveal a reduction in dimensionality, which leads to improved scalability. Initially, agents receive the state with the coverage time value, the total number of nodes connected to the RSU per category, and the category (service type). Using this information, agents can learn an optimal policy for selecting the appropriate waiting time for transmission. In a multi-agent scenario, with different categories assigned to AVs for data transfer, we can consider distinct agent networks. We define the network of all agents as $\Phi$, with state $S = \{S_j, T_v, C, T_{cv}\}$, and the network of agents per category as $\Phi(c)$. Analyzing the state for the network $\Phi$ especially by subdividing the network per category, it is found that the agent's state space, the category is the same, which becomes redundant. Thereby, it could be established that the state space is reduced by one variable (Eq. \eqref{eq:multi_agent_equation2}). Therefore, for each agent, the state becomes $S = \{S_j, T_v, T_{cv}\}$. This highlights the fact that a multi-agent system can effectively reduce the state space. For instance, if we provide sample values for each variable in the state, we could quantify how much the state space is optimized. Given the length of $S_j$ of 5, $|T_v|$ of 100, $|T_{cv}|$ also 100, and $|C|$ of 4 (per our scenario). Then, the size of the state space with 4 variables is $|S_j|\times|T_v|\times|T_{cv}|\times|C| = 200000$, and for 3 variables is $|S_j|\times|T_v|\times|T_{cv}| = 50000$, which is $75\%$ decrease.

\subsubsection{Actions}
A set of actions can be mapped by Eq. \eqref{eq:map_actions}, where  $\mathcal{A} = \{0,..,k\}$, and $k$ is the maximum number of actions. The equation converts the discrete value $a$ into a continuous value by considering the maximum waiting time per category $w_{max}(c)$, which is described in our previous work \cite{our_sojourn_single_agent}.
\begin{equation}
    w(c) = a \cdot \left(\frac{w_{max}(c)}{|\mathcal{A}|}\right)
    \label{eq:map_actions}
\end{equation}

\begin{figure}[htbp]
\centerline{\includegraphics[width=\columnwidth,height=5cm,keepaspectratio]{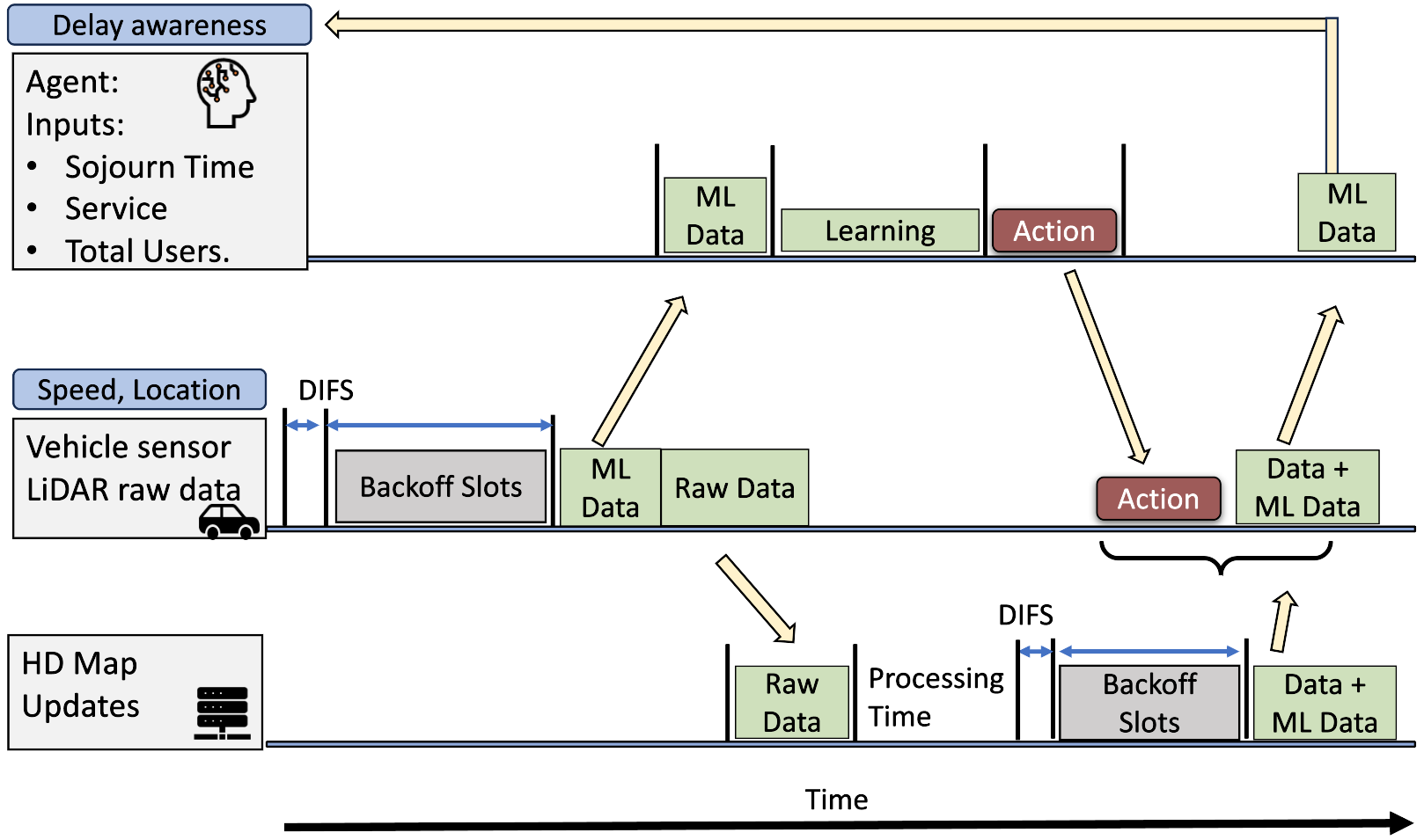}}
\caption{The distributed approach of data flow in the time domain from agent to vehicle.}
\label{fig:data_flow_figure}
\end{figure}

\subsubsection{Algorithm} 
This work builds upon our prior research in single-agent coverage awareness and RL \cite{our_sojourn_single_agent}, where we propose extending the foundational Algorithm 3 to accommodate a distributed multi-agent framework. The details are presented in Algorithm \ref{algo:q_learning} of this paper. The solution is achieved by adding a for loop, allowing the acquisition of data and learning of policy for each agent, described in Line 3 of Algorithm \ref{algo:q_learning}. Besides, in Line 10, the AV agent calculates the reward based on individual or overall network performance, which includes latency and throughput information provided by the edge server, as stated in the test cases in the next section. In Line 8 of the same algorithm, the action is obtained by the subroutine choose\_action, which is described in Algorithm \ref{algo:checkActiveUsers} \cite{our_sojourn_single_agent} that includes the e-greedy procedure and the Eq. \eqref{eq:map_actions}. A visual description of the data flow of the data distribution is depicted in Fig. \ref{fig:data_flow_figure}. It can be observed form the figure that the vehicle sends the corresponding data to the agent, an HD Map server. The agent uses this information for learning purposes. After that, the edge server provides the network parameters $T_v$, $T_{cv}$, $\mathcal{L}$, and $\mathcal{R}$ to generate the state and calculate the reward function.

\begin{algorithm}
\caption{Q-Learning}
\label{algo:q_learning}
\begin{algorithmic}[1]
\STATE Initialize the agent with epsilon $\epsilon
$ , number of actions $k$.
\
\FOR {number\_of\_episodes}
\FOR {each agent} 
  \WHILE{STATUS}
    \STATE Sojourn Time = $\frac{(area/2) - d}{speed}$, then a discrete value is selected $S_j =\{0,1,2,3,4\}$ as described in \cite{our_sojourn_single_agent}.
    \STATE \text{agent} $\leftarrow$ $T_v$, $T_{cv}$, $\mathcal{L}$, and $\mathcal{R}$
    \STATE agent $\rightarrow$ $\mathcal{S} = \{S_j , T_v , C, T_{cv}\}$
    \STATE $a$ $\leftarrow$ choose\_action(S)
    \STATE AV $\leftarrow$ $a$
    \STATE $r_t$ $\leftarrow$ based on the overall latency and throughput, according to the utility function (\ref{eq:utility_function_penalty}) 
    \STATE agents $\leftarrow$ next state, $r_t$, STATUS
    \STATE update $Q(s,a)$ by using Eq. (\ref{eq_q_learning_2})
  \ENDWHILE
\ENDFOR
\ENDFOR
\STATE \textbf{Finish the subroutine}
\end{algorithmic}
\end{algorithm}

\begin{algorithm}
\caption{CheckActiveUsers Subroutine \cite{our_sojourn_single_agent}}
\label{algo:checkActiveUsers}
\begin{algorithmic}[1]
\STATE Find user
\STATE Get counter
\IF{User found} 
    \STATE decrease counter by 1
    \IF{counter == 0}
    \STATE delete user from map TotalActiveUsers
    \ENDIF
\ENDIF
\end{algorithmic}
\end{algorithm}

\section{Performance Analysis}
\begin{figure}[htbp]
\centerline{\includegraphics[width=\columnwidth,keepaspectratio]{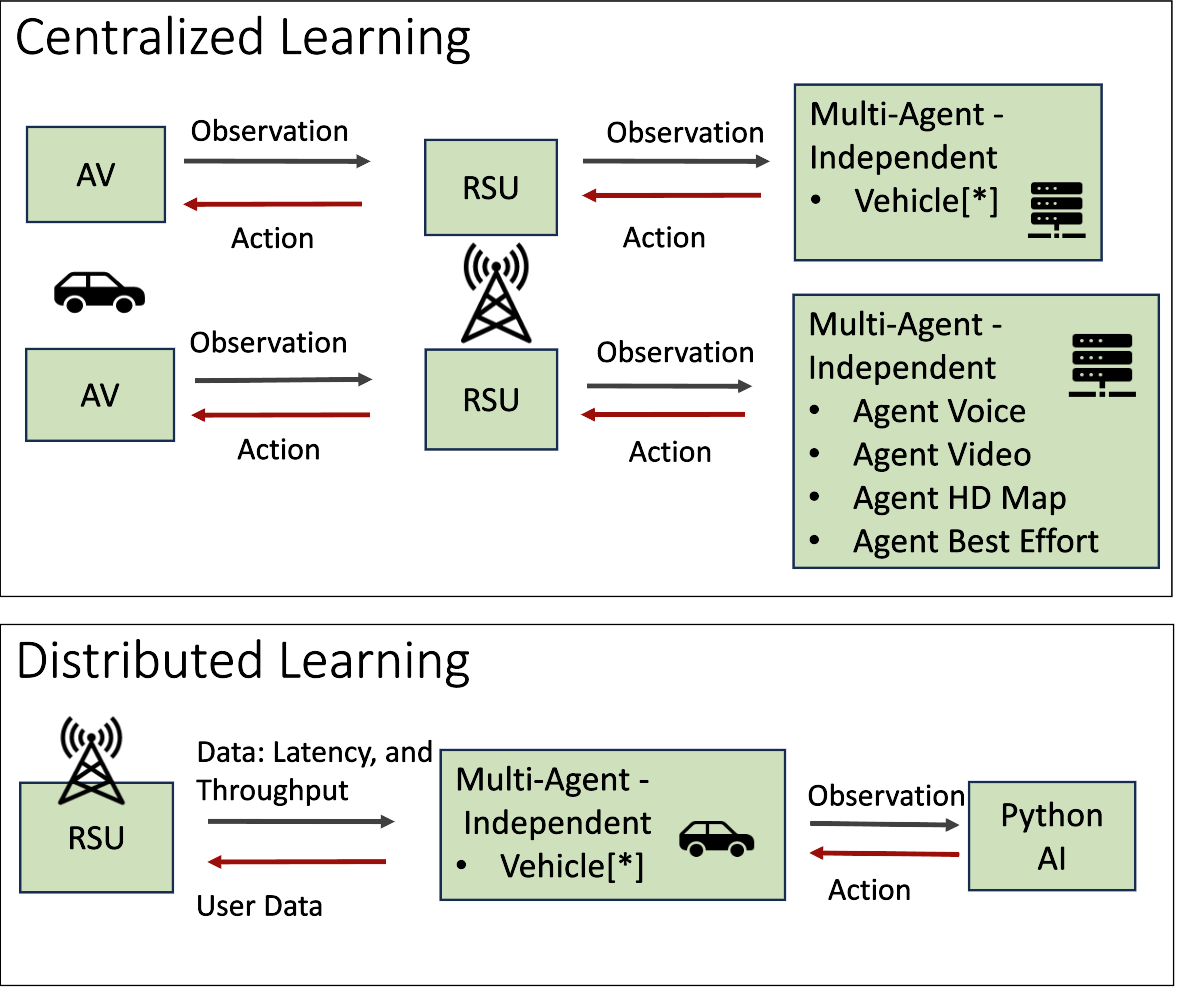}}
\caption{Centralized and distributed learning architecture.}
\label{fig:centralize_distributed}
\end{figure}

To assess the scalability of a single-agent solution in a multi-agent setup and determine the optimal number of agents for a VANET environment, four test case scenarios are devised to verify the feasibility.
In the first scenario, the simulation is focused on centralized learning and reward function calculation, where the AVs will send the information to the edge server. There is a dedicated entity for each of the agents in the edge server, as shown in Fig. \ref{fig:centralize_distributed}. Test cases two and three involve different numbers of agents. Finally, test case 4 involves centralized and distributed learning.

\subsection{\textbf{Test Case 1}: Exploring Reward Calculation Strategies in Multi-Agent Systems: Node-specific vs. Overall Application Analysis}
For this test case, we aim to explore different reward calculation strategies in multi-agent systems. We will compare the results of calculating the reward per node (AV) versus calculating the reward for the overall application. In the first simulation, each agent will calculate its reward based on its current latency and throughput as follows $\text{each } v \leftarrow c\text{, where } \mathcal{L}th_{v} = \mathcal{L}_{max}(c)\text{, and } \mathcal{R}th_{v} = \mathcal{R}_{min}(c)$, and the latency and throughput are $\mathcal{L}_v$, and $\mathcal{R}_v$. Thus, the reward equation becomes,
\begin{equation}
    \begin{split}
    r_{v} = \alpha_1\frac{\mathcal{R}_v}{\mathcal{R}th_{v}}- \alpha_2\frac{\mathcal{L}_v}{\mathcal{L}th_{v}} + \mathcal{F}
    \label{eq:reward_per_vehicle}
    \end{split}
\end{equation}
where $\mathcal{L}th_{v}$, and $\mathcal{R}th_{v}$, are the thresholds. The thresholds are described in Table \ref{tab:maxi_actions} which are the desired requirements for each of the services. In the second simulation, we will focus on calculating the reward using average latency and throughput per service, as measured by the edge server. Thus, the latency and throughput are denoted as $\mathcal{\overline{L}}$, and $\mathcal{\overline{R}}$ respectively. Thus, the reward equation becomes,

\begin{equation}
    \begin{split}
    r_{v} = \alpha_1\frac{\mathcal{\overline{R}}}{\mathcal{R}th_{v}}- \alpha_2\frac{\mathcal{\overline{L}}}{\mathcal{L}th_{v}} + \mathcal{F}
    \label{eq:reward_overall}
    \end{split}
\end{equation}

\subsection{\textbf{Test Case 2}:Enhancing Service-Specific Performance: Optimal Agent Allocation per Service Type}
\begin{figure}[htbp]
\centerline{\includegraphics[width=6cm,height=3cm,keepaspectratio]{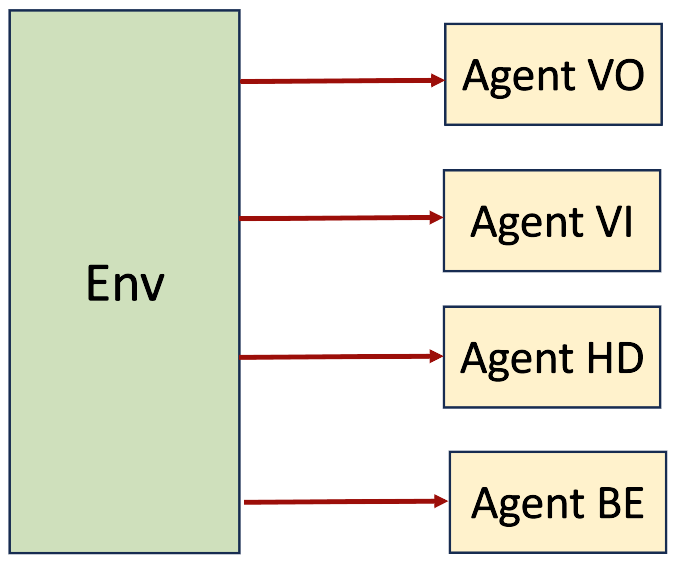}}
\caption{Diagram of the agents per service.}
\label{fig:agents_per_service}
\end{figure}

This approach enables us to delineate the responsibilities of each agent and determine the necessary quantity. Under this arrangement, the number of agents corresponds directly to the number of service categories. Transitioning from a single agent managing actions for all vehicles to assigning a dedicated agent for each service. It reduces the scope of responsibility per agent, thereby streamlining the local environment. Consequently, this segregation of agents per service has the potential to optimize individual agent performance. For a visual representation, please refer to Fig. \ref{fig:agents_per_service}. Let set each category $c \in \mathcal{C}$ as an agent. Each agent c will calculate the reward as follows:

\begin{equation}
    \begin{split}
    r_{c} = \alpha_1\frac{\mathcal{\overline{R}}}{\mathcal{R}th_{c}}- \alpha_2\frac{\mathcal{\overline{L}}}{\mathcal{L}th_{c}} + \mathcal{F}
    \label{eq:reward_per_category}
    \end{split}
\end{equation}

\subsection{\textbf{Test Case 3}: Exploring Performance Variations: Agent Allocation per Service Type vs. Vehicles as Multi-Agent Entities}
\begin{figure}[htbp]
\centerline{\includegraphics[width=6cm,height=3cm,keepaspectratio]{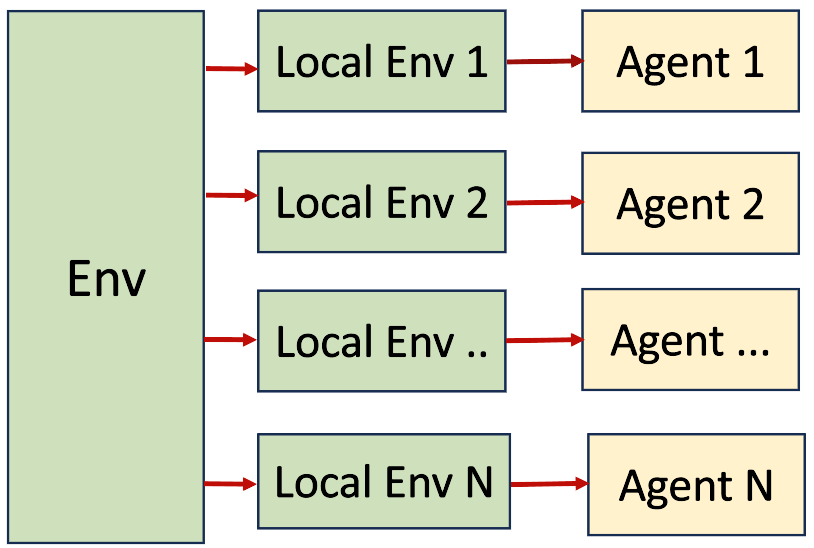}}
\caption{Diagram of the agents per autonomous vehicle.}
\label{fig:agents_per_vehicle}
\end{figure}

We aim to evaluate if different vehicles, acting as agents, display varied performance levels with this test case. By using each vehicle as an agent, the state space is reduced due to a smaller local environment when compared to the previous test case (refer to Figure \ref{fig:agents_per_vehicle}). The reward function calculation follows Eq. \eqref{eq:reward_overall}.

\subsection{\textbf{Test Case 4}: Centralized vs. Distributed Learning for AVs as Agents}
For the final assessment, each of the autonomous vehicles (AVs) will be chosen as agents to be tested in both centralized and distributed learning approaches. This will help to determine the costs in terms of latency and throughput of the overhead generated in the centralized approach by sending information to the edge server for learning.
If we consider the straightforward calculation for queuing delay \cite{performance_analysis_HDMAP}, as depicted in Eq. \eqref{eq:eq_queue_delay}, latency is bound to increase in centralized learning mode. This is a direct result of the increased number of packets because of data sharing among vehicles and the transmission of actions from agents to AVs, described as below: 

\begin{equation}
\text{Queue}_{\text{delay}}=  \sum_{k=1}^{K}\Biggr[ f_{t}[n] \Biggr]*bf_{t}   \label{eq:eq_queue_delay}
\end{equation}
where the variables $K$, $bf_{t}$, and $f_{t}$ represent the number of packets on the queue, the backoff time, and the time required for packet transmission, respectively, depending on its size.

\section{Simulation Setup}\label{simu}
The current standard IEEE802.11p with a 10MHz bandwidth for the simulation is utilized in our simulation setup. To create a realistic VANET environment, we chose OMNet++ version 6 \cite{omnetppOMNeTDiscrete} and the INET \cite{omnetppINETFramework} library, due to the Physical and Mac protocol stack for IEEE802.11p. To generate traffic flow, we use SUMO \cite{SUMO} and the OMNet++ module veins \cite{veins}. Additionally, a research group solution \cite{veinsgym} to connect OMNet++ and Python for the RL is employed and adapted to our simulation.

\begin{table}[ht]
\caption{Simulation Parameters}
\begin{center}
\begin{tabular}{|c|c|}
\hline
\textbf{\textit{Parameter}} & \textbf{\textit{Value}} \\
\hline
Tile Dimension & 300x100 meters  \\
\hline
Episodes & 50  \\
\hline
Episode duration & 250s  \\
\hline
$\epsilon-greedy$ & 0.2  \\
\hline
Discount Factor $\gamma $& 0.99  \\
\hline
Learning Rate $\epsilon $& 0.1  \\
\hline
$\alpha_1 $, $\alpha_2 $ & 0.3 and 0.7 respectively  \\
\hline
Simulation time & 250s  \\
\hline
Vehicle Density & varies according traffic flow  \\
\hline
Coverage Area & 200m  \\
\hline
Vehicle max speed & 17m/s  \\
\hline
Vehicle Acceleration  & {2.6 m}/{$s^2$} Average value of petrol cars\cite{acceleration_deceleration_petrol_car}. \\
\hline
Vehicle Deceleration  & {4.5 m}/{$s^2$} Average value of petrol cars\cite{acceleration_deceleration_petrol_car}.  \\
\hline
Tx Power & 200 mW  \\
\hline
Frequency & 5.9 GHz  \\
\hline
Bandwidth & 10MHz  \\
\hline
Best-effort data rate & 28Mbps  \\
\hline
HD Map data rate & 4Mbps \\
\hline
Video data rate & 5Mbps  \\
\hline
Voice data rate & 100kbps  \\
\hline
TXOP limit & Disabled as per standard  \\
\hline
\end{tabular}
\label{table:simulation_parameters}
\end{center}
\end{table}
The dataset from Traffic and Accident Data Unit/North East Regional Road Safety Resource is used \cite{dataset} in our experiment, with an entry interval time per vehicle of $0.66$s, following our previous work \cite{our_sojourn_single_agent}. Vehicles will enter and exit the environment during the simulation.
Upon entering the simulation, each vehicle will be randomly assigned a category. The data rate per category can be found in Table \ref{table:simulation_parameters}.
Finally, to ensure a valid comparison between single and multi-agent approaches, and to demonstrate scalability, we have maintained consistent values for maximum waiting time (actions), thresholds, and rewards. Please refer to Table \ref{tab:maxi_actions} for further details.

\begin{table}[h]
\centering
\caption{Maximum waiting time and thresholds per category \cite{our_sojourn_single_agent}}
\begin{tabular}{|c|c|c|c|}
   \hline
   Service & max wt & $\mathcal{R}$ Threshold & $\mathcal{L}$ Threshold\\
   \hline
   VO  & 0.92s & 100kpbs \cite{dataRate_cisco_voice_szigeti2005end} &  150 ms \cite{cisco_2} \\
   \hline
   VI  & 2s & 1.25Mbps \cite{dataRateVideo_googleYouTube} & 100 ms \cite{dataRate_cisco_voice_szigeti2005end}\\
   \hline
   HD Map  & 2s & 1.25Mbps & 100 ms \cite{5gaa_delay_2}\\
   \hline
   BE  & 8s & 1.0Mbps  & 1000 ms \\
   \hline
   \multicolumn{2}{l}{wt = Waiting time. } \\
\end{tabular}
\label{tab:maxi_actions}
\end{table}

\section{Experimental Results and Analysis}\label{analysis}
This section presents the results and discussions stemming from a comprehensive comparison between the single-agent and multi-agent approaches. The comparison is measured with key performance indicators (KPIs) such as latency, throughput, and fairness. Latency is defined as the time difference between the creation and reception of a packet. Throughput is quantified by the total number of packets received within a designated time interval. To assess fairness, Jain's fairness index formula is employed \cite{fairness_jain}, as shown in Eq. (\ref{eq:fairness}).

\begin{equation}
    J(x_1,x_2,...,x_n) = \frac{(\sum_{i=1}^n x_i)^2}{n \cdot \sum_{i=1}^n x_{i}^2}
    \label{eq:fairness}
\end{equation}

The results are displayed in a Cumulative Distribution Function (CDF) graph for a comprehensive view of latency and throughput distribution. It facilitates understanding of the entire range of latency and throughput and enables straightforward comparison across different network scenarios.

\subsection{\textbf{Test Case 1}: Exploring Reward Calculation Strategies in Multi-Agent Systems: Node-specific vs. Overall Application Analysis}
\begin{figure}[ht]
\centerline{\includegraphics[width=\columnwidth, keepaspectratio]{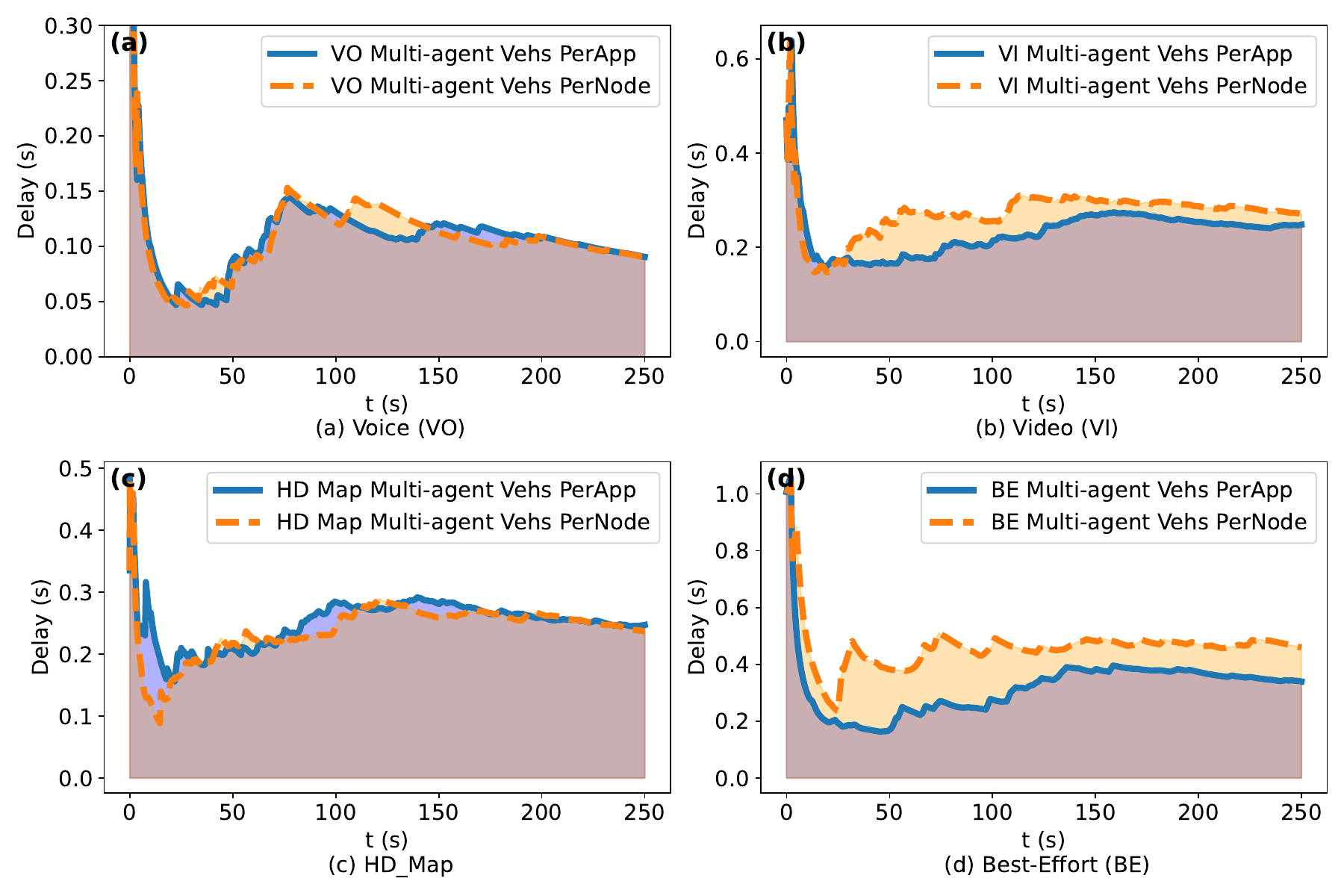}}
\caption{Latency comparison between node-specific reward calculation and the overall average reward per service.}
\label{fig:latency_per_node_app}
\end{figure}

\paragraph{Latency}
From the results of VO in Fig. \ref{fig:latency_per_node_app}(a), and HD Map in Fig. \ref{fig:latency_per_node_app}(c), there is no significant difference in whether the reward is calculated using the average latency, and throughput per application or node-specific. The latency percentage difference is $0.43\%$ for VO and $4.3\%$ for HD Map. Concerning VI and BE in Fig. \ref{fig:latency_per_node_app}(b), and Fig. \ref{fig:latency_per_node_app}(d), there is a difference in latency of $8.6\%$ and $29.8\%$, respectively.

\begin{figure}[ht]
\centerline{\includegraphics[width=\columnwidth,keepaspectratio]{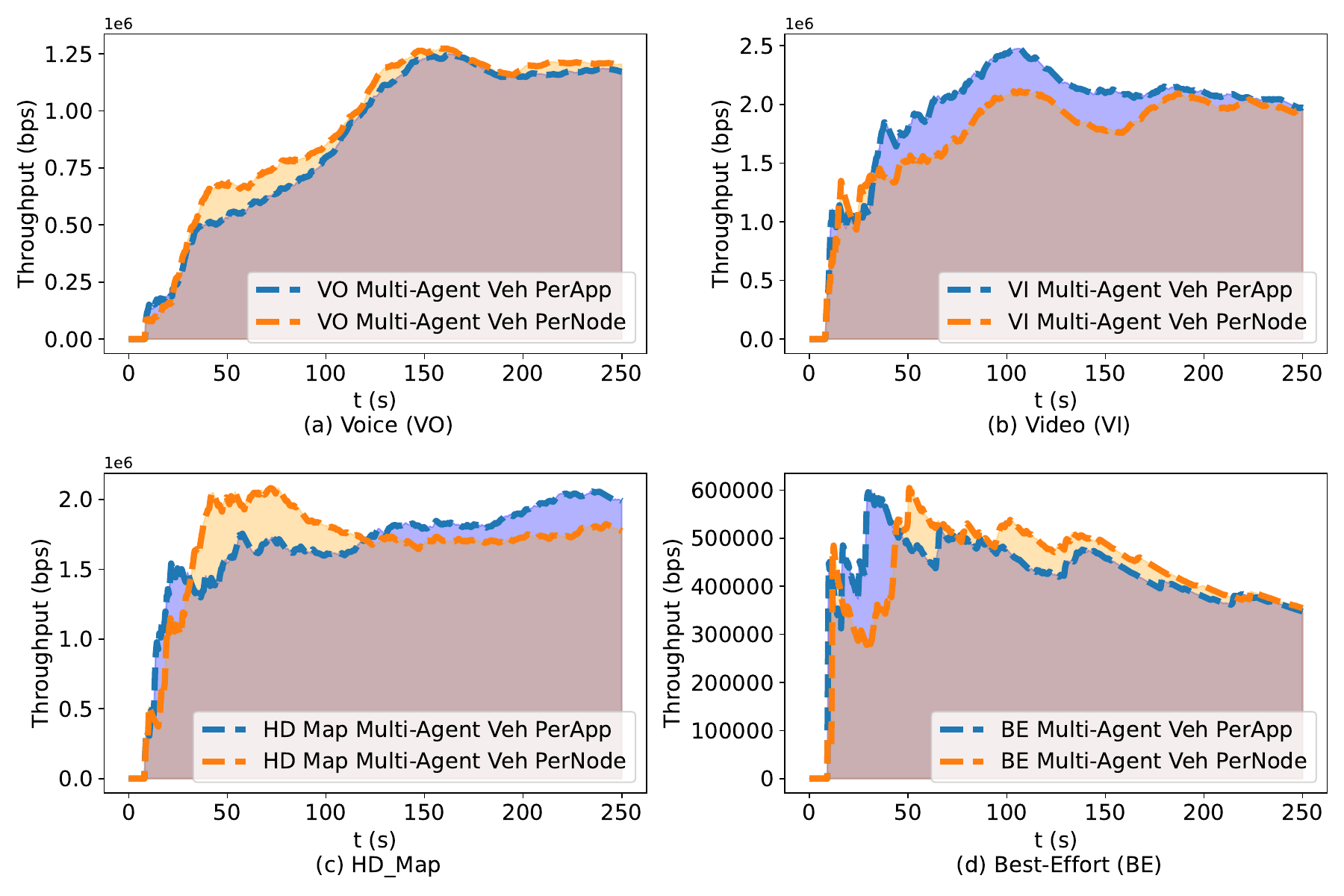}}
\caption{Throughput comparison between node-specific reward calculation and the overall average reward per service.}
\label{fig:throughput_per_node_app}
\end{figure}

\paragraph{Throughput}
For the throughput results in Fig. \ref{fig:throughput_per_node_app}, it is observed a small percentage increase while calculating the reward with the overall latency and throughput per application compared with node-specific for VI, and BE. The percentage increase is $2.9\%$ for VI, and $1.9\%$ for BE, as described in Fig. \ref{fig:throughput_per_node_app}(b), and Fig. \ref{fig:throughput_per_node_app}(d), respectively. For VO and HD Map, the average throughput does not vary significantly.

These results demonstrate that the optimal policy varies depending on the perception and approach taken to address the problem. Moreover, this reveals that the solution remains robust even when using different reward calculations, as long as the agents utilize the same reward function. To generate the possible lowest latency, it is clear that the overall average latency per application is more efficient for all four types of services.

\subsection{\textbf{Test Cases 2 and 3}: An agent per service type, and per vehicle}
When examining test cases 2 and 3, it becomes evident that utilizing a multi-agent approach yields better results when compared to a single agent. To conduct this comparison, we calculate the reward with the average per-application approach, as per the results in test case 1. The two distinct types of agents are labeled as multi-agent ACs and multi-agent Vehs, which correspond to service type and per vehicle, respectively.

\begin{figure}[ht]
\centerline{\includegraphics[width=\columnwidth,keepaspectratio]{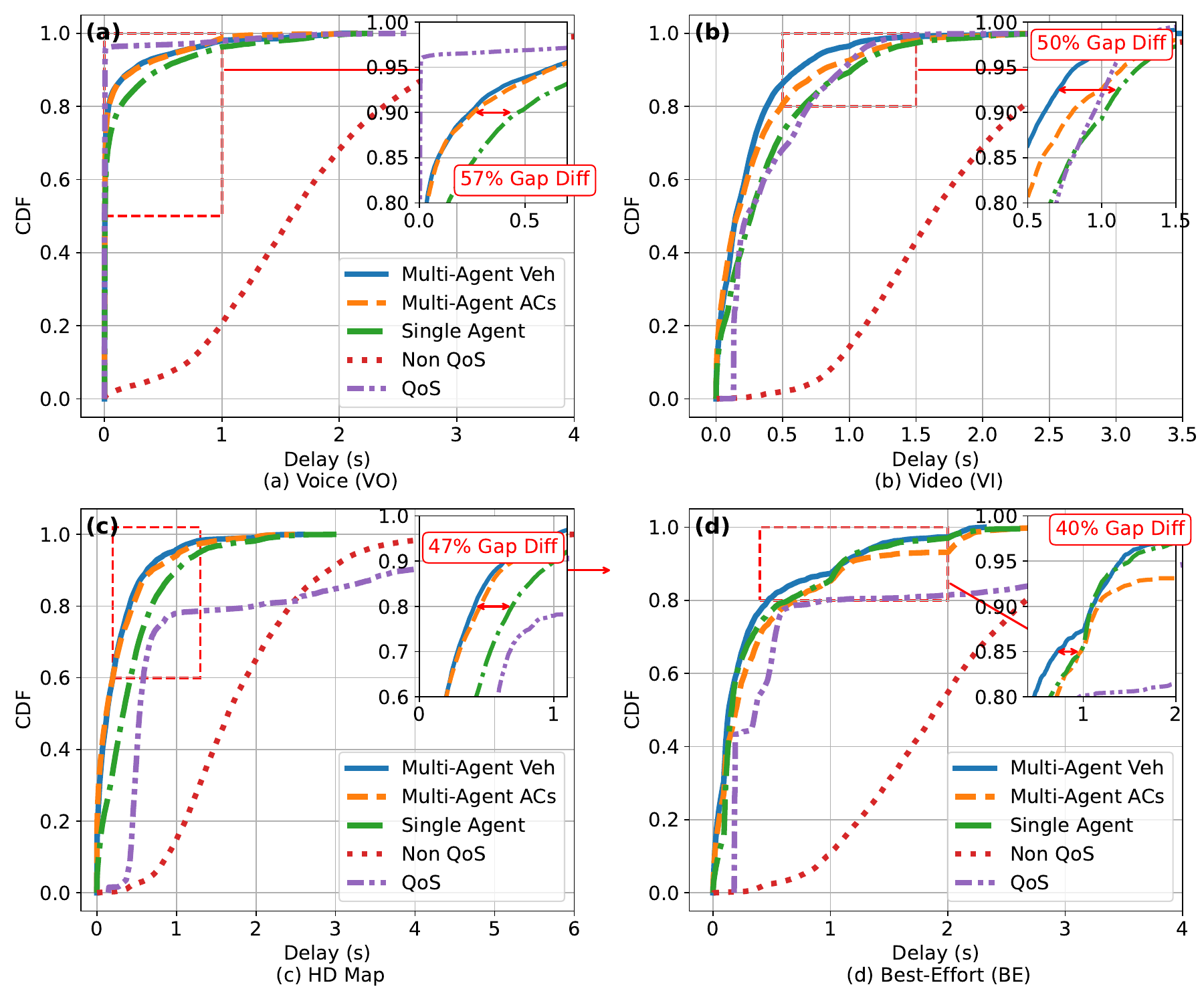}}
\caption{CDF Latency comparison between standard (Non QoS), standard QoS (EDCA), single agent, and multi-agent.}
\label{fig:cdf_latency}
\end{figure}

\begin{figure}[ht]
\centerline{\includegraphics[width=\columnwidth,keepaspectratio]{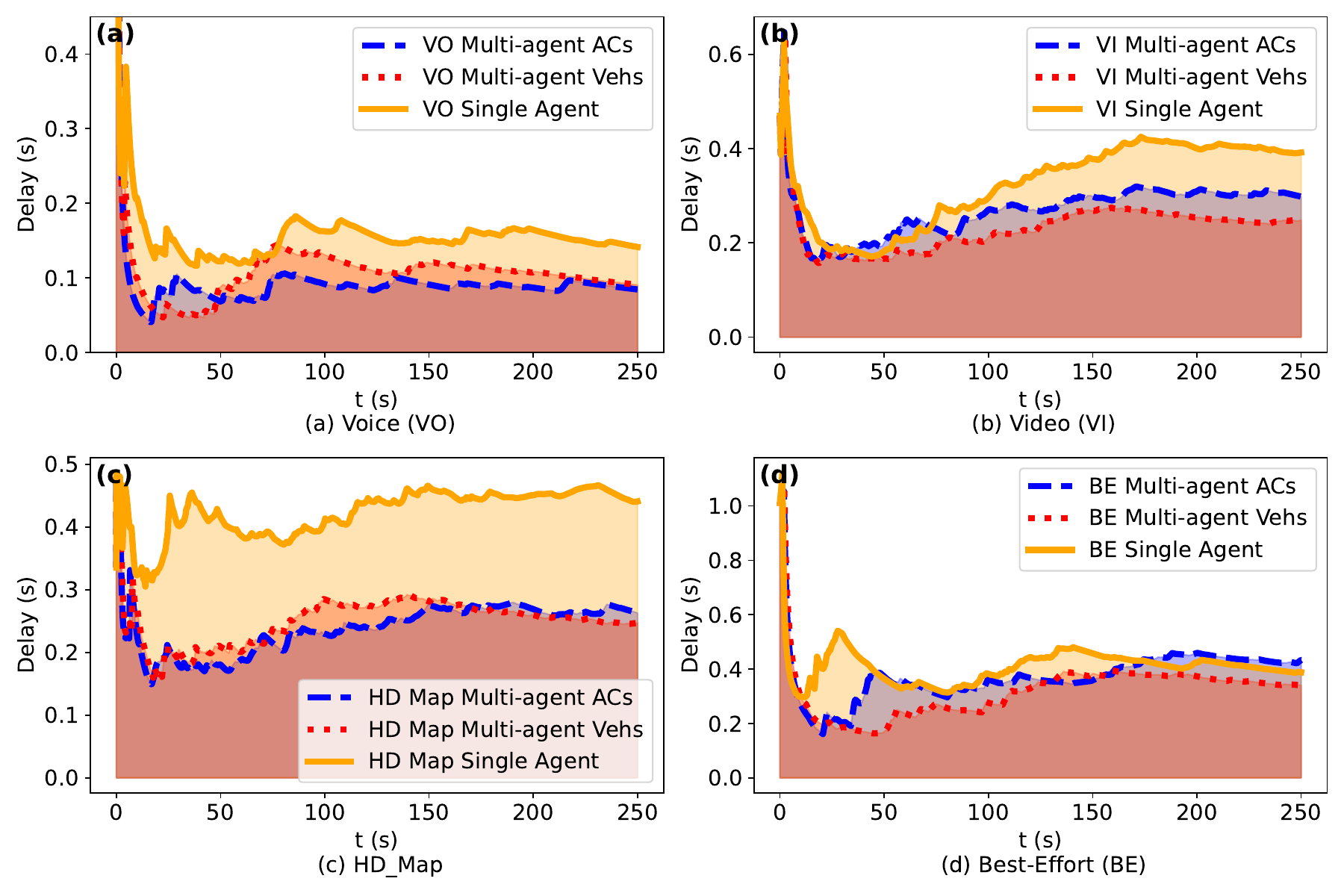}}
\caption{Latency comparison between single and multi-agent approach.}
\label{fig:time_domain_latency}
\end{figure}

\paragraph{Latency}
In Fig. \ref{fig:cdf_latency}, it is noticed that the multi-agent solution outperforms the single agent for all types of services. For the CDF latency of VO in Fig. \ref{fig:cdf_latency}(a), the multi-agent approach was able to reach a lower delay with a gap difference of $57\%$ compared to a single agent between $80\%$ and $100\%$ in the CDF. The graph also illustrates that the $90\%$ of the simulation values are below $0.25s$ for the multi-agent approach compared with $0.5s$ for a single agent. Furthermore, Fig. \ref{fig:time_domain_latency}(a) shows that during the simulation, the multi-agent approach maintained latency with an average of $0.0844$s for multi-agent ACs, and $0.0904$s for multi-agent Vehs, which is $40.4\%$ and $36\%$ lower than the latencies with a single agent respectively.

For VI in Fig. \ref{fig:cdf_latency}(b), the multi-agent Vehs achieves lower latency compared to the latencies using other approaches. The multi-agent Vehs has a gap difference of $50\%$ having values below $0.7s$ while a single agent performs at $1.2s$ within the $80\%$-$100\%$ range in the CDF. Furthermore, Fig. \ref{fig:time_domain_latency}(b) highlights an improvement in the delay of around $23.8\%$ for multi-agent ACs and $36\%$ for multi-agent Vehs.

For HD Map, Fig. \ref{fig:cdf_latency}(c), both the multi-agent ACs and Vehs have a gap difference of $47\%$ to the single agent. The improvement is notable in Fig. \ref{fig:time_domain_latency}(c), with a reduction in the latency around $40.2\%$ for multi-agent ACs and $43.8\%$ for multi-agent Vehs compared with the single-agent approach.

For the lowest priority category, BE, as shown in Fig. \ref{fig:cdf_latency}(d), the three solutions display similar behaviour. Only multi-agent Vehs showed an improvement in part of the CDF graph with a gap difference of $40\%$. This corresponds to an improvement of $12.46\%$, as described in Fig. \ref{fig:time_domain_latency}(d).

\begin{figure}[ht]
\centerline{\includegraphics[width=\columnwidth,keepaspectratio]{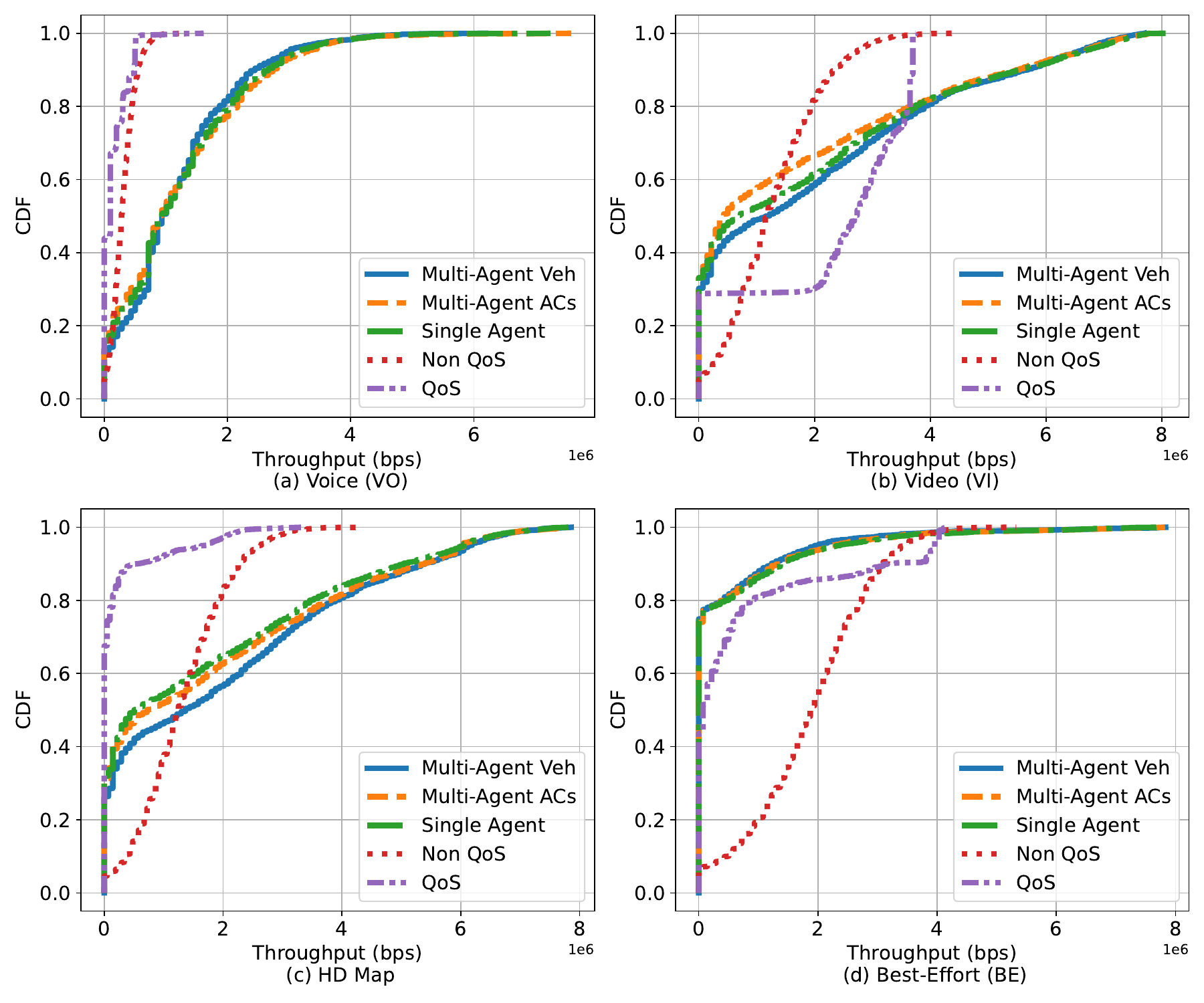}}
\caption{CDF Throughput comparison between standard (Non QoS), standard QoS (EDCA), single agent, and multi-agent.}
\label{fig:cdf_throughput}
\end{figure}

\begin{figure}[ht]
\centerline{\includegraphics[width=\columnwidth,keepaspectratio]{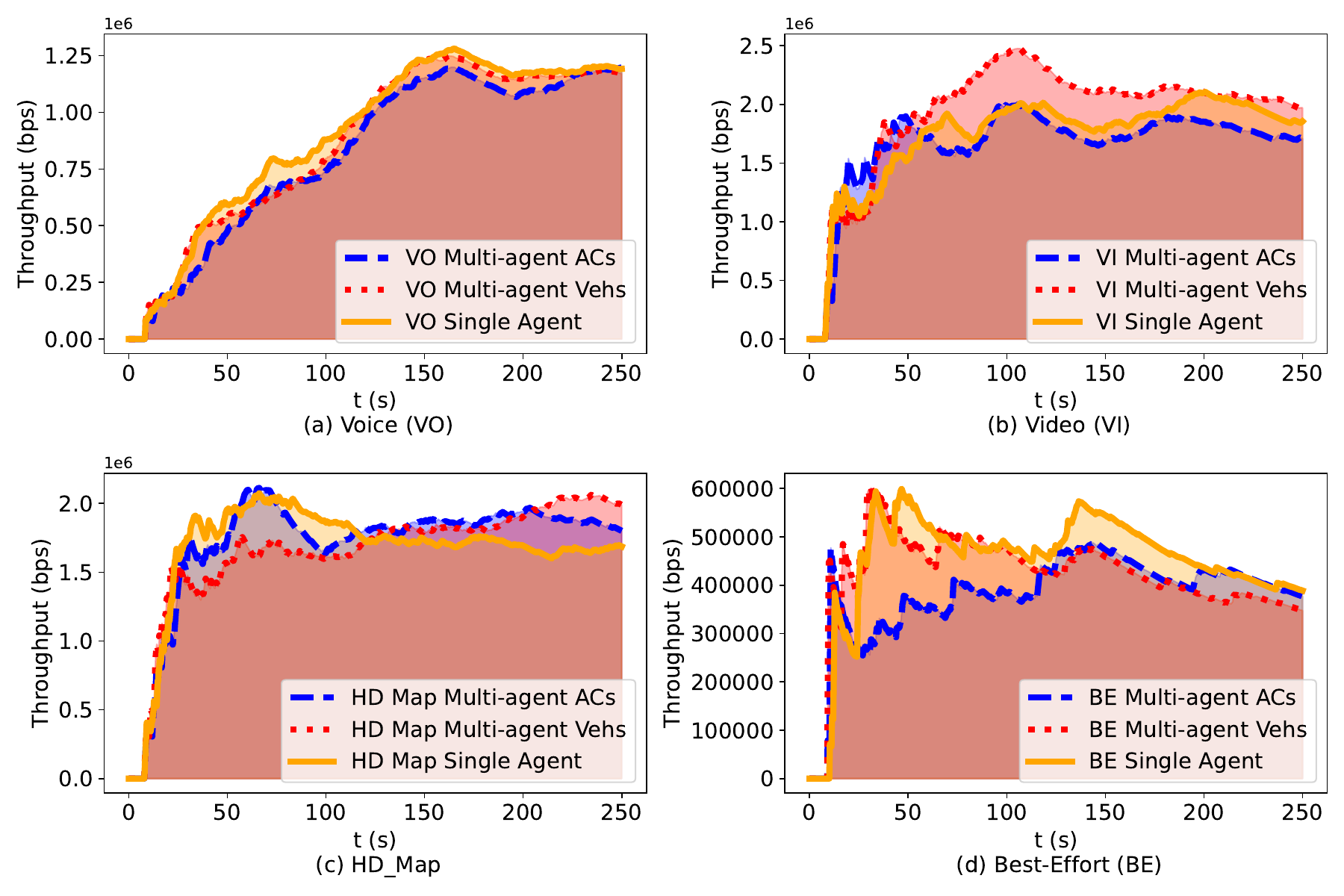}}
\caption{Throughput comparison between single and multi-agent approaches.}
\label{fig:time_domain_throughput}
\end{figure}

\paragraph{Throughput}
It can be observed from the CDF throughput in Fig. \ref{fig:cdf_throughput} that the single and multi-agent systems can provide higher throughput for VO, VI, and HD Map services. Regarding BE, the throughput is reduced as it is the lowest priority, which is the expected outcome. This reveals how the solution that operates in the application layer provides service priorities. Additionally, it is observed that the lines of multi-agent scenarios are located rightmost in the CDF for the services VO, VI, HD Map in  Figs. \ref{fig:cdf_throughput}(a), \ref{fig:cdf_throughput}(b), and \ref{fig:cdf_throughput}(c), respectively. These observations indicate that the RL algorithms employed in the context of multi-agent environments achieve higher throughputs. On the contrary, for BE in Fig. \ref{fig:time_domain_throughput}(d), we noticed a marginal decline in throughput, which is in line with our objective of maintaining this category as a low priority.

In more details of VO service, it is observed from Fig. \ref{fig:time_domain_throughput}(a) that there is a slight difference in throughput between both single and multi-agent approaches. Referring to VI in Fig. \ref{fig:time_domain_throughput}(b), the multi-agent Vehs generated higher throughput compared to multi-agent ACs and single agent with a percentage increase of $6.19\%$, and $15.3\%$ respectively. A similar behaviour is recognised in Fig. \ref{fig:cdf_throughput}(c), and Fig. \ref{fig:time_domain_throughput}(c) for HD Map. In this service, the multi-agent Vehs generated a higher throughput at a certain time of the simulation with an average increase of $17\%$ compared to a single agent, and $10.4\%$ compared to the throughput of multi-agent ACs. 

\subsection{\textbf{Test Case 4}: AVs as agents, centralized vs. distributed learning}
In this test case, each vehicle was considered an agent, the main difference relates to where the processing occurs. In centralized learning, the agents are located in the edge server, and the information is shared by the AVs to the agents accordingly. For distributed learning, the vehicle collects the data and the learning occurs directly within the AV. 

\begin{figure}[h]
\centerline{\includegraphics[width=\columnwidth,keepaspectratio]{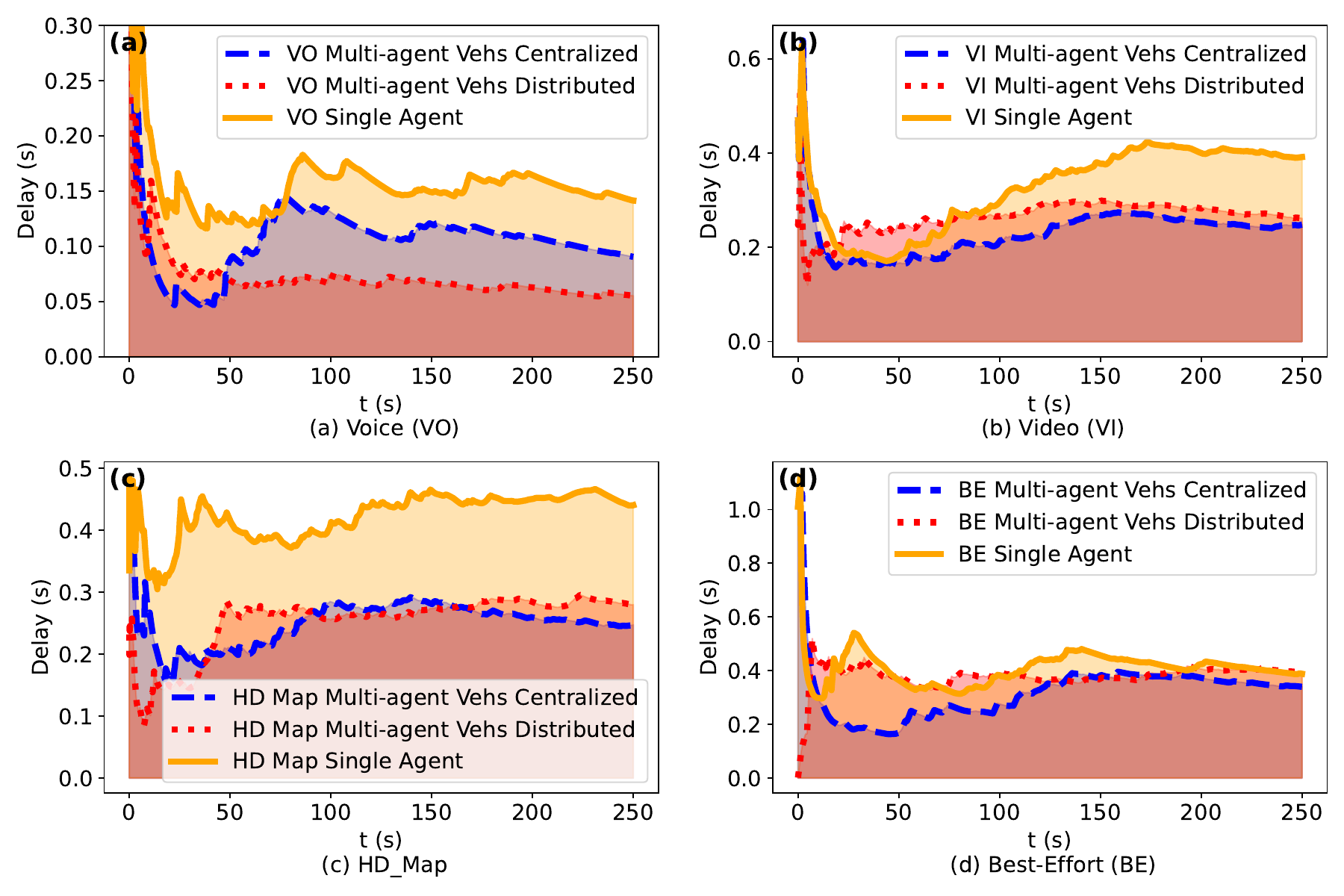}}
\caption{Latency comparison between centralized and distributed learning approaches.}
\label{fig:centra_distr_latency}
\end{figure}

\paragraph{Latency}
For VO service and distributed learning in Fig. \ref{fig:centra_distr_latency}(a), the latency is reduced by $32.7\%$ compared to centralized learning, and $54\%$ compared to a single agent. For VI in Fig. \ref{fig:centra_distr_latency}(b), the latency difference between the centralized and distributed approaches is $11.5\%$. For HD Map category Fig. \ref{fig:centra_distr_latency}(c), and BE Fig. \ref{fig:centra_distr_latency}(d) the latency difference is $1.15\%$, and $17.5\%$ between distributed and centralized learning approach, respectively.

\begin{figure}[h]
\centerline{\includegraphics[width=\columnwidth,keepaspectratio]{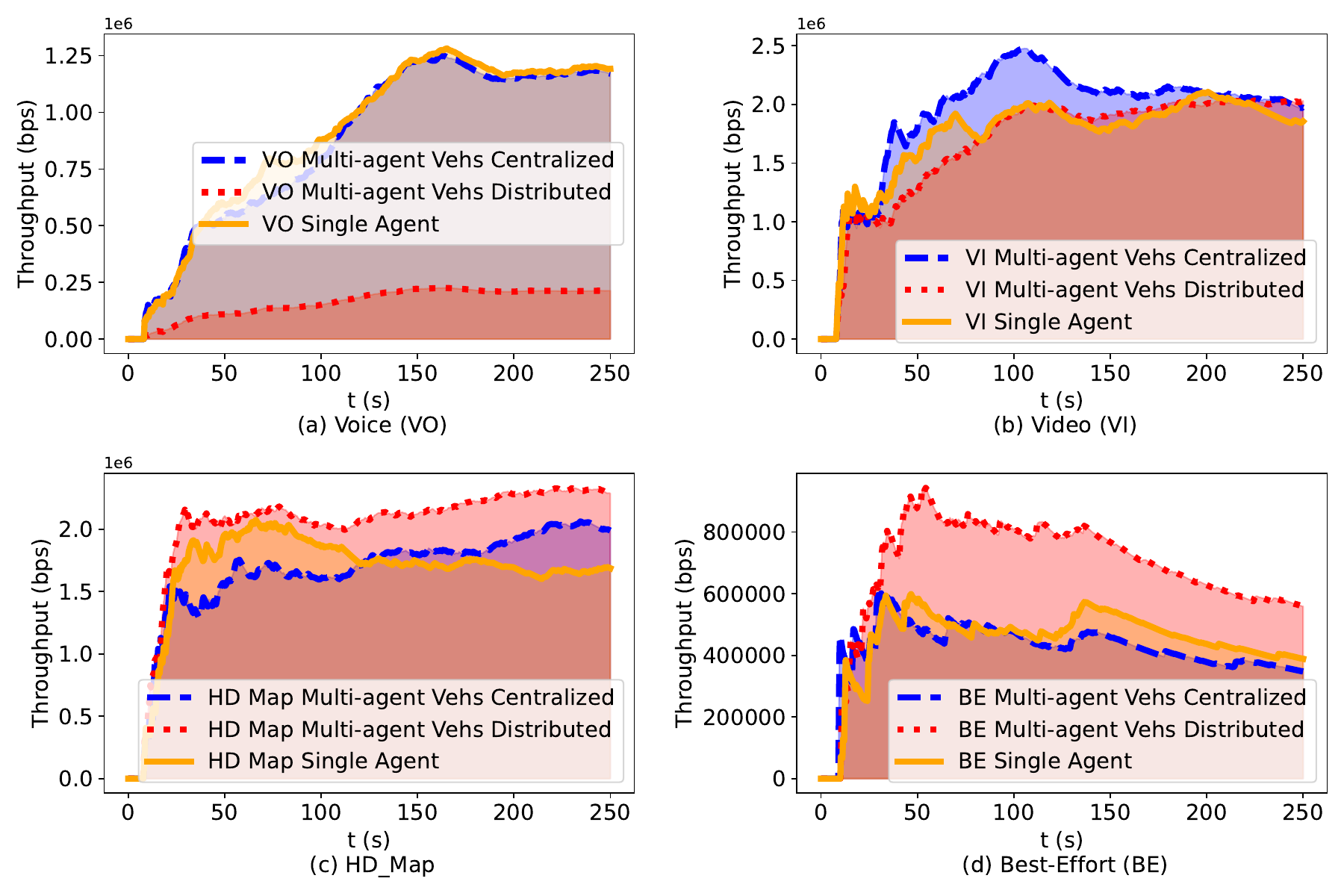}}
\caption{Throughput comparison between centralized and distributed learning approaches.}
\label{fig:centra_distr_throughput}
\end{figure}

\paragraph{Throughput}
The throughput results obtained by the distributed learning approach have shown a higher priority control. This is observed in Fig. \ref{fig:centra_distr_throughput}(a) where the throughput is maintained close to the threshold of $100$ kbps. Similar results were exhibited in Fig. \ref{fig:centra_distr_throughput}(b) for VI where the average throughput is about the threshold $1.25$Mbps. Besides, for the HD Map service, the distributed scheme yielded higher throughput with a percentage increase of $22\%$, as perceived in Fig. \ref{fig:centra_distr_throughput}(c). Lastly, it is observed in Fig. \ref{fig:centra_distr_throughput}(d) that there is a staggering increase of $62\%$ in BE.

\begin{figure}[ht]
\centerline{\includegraphics[width=\columnwidth,keepaspectratio]{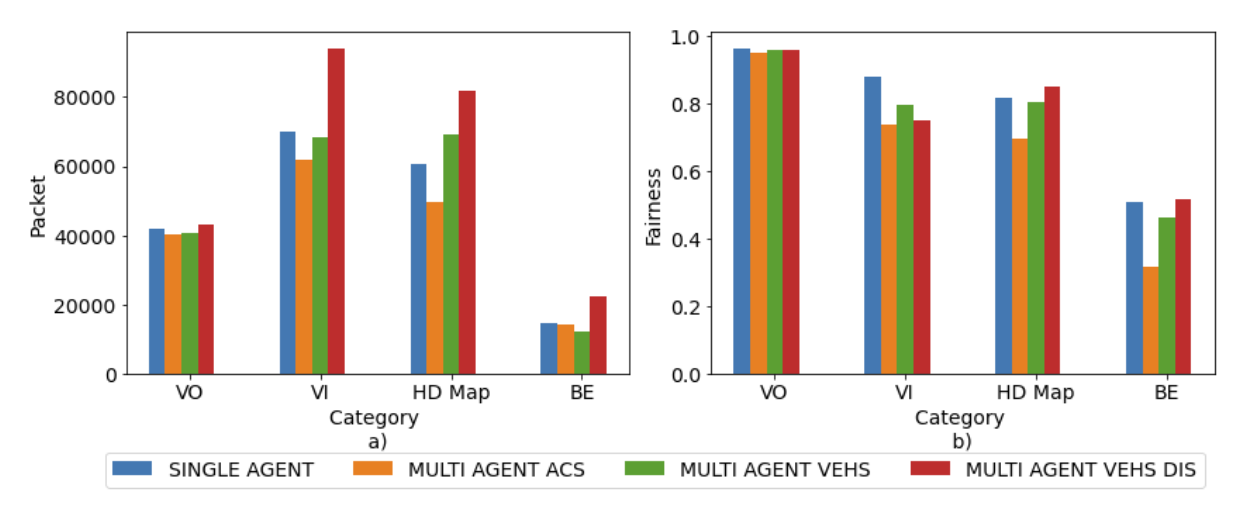}}
\caption{Packet received and fairness comparison between single agent and multi-agent.}
\label{fig:overall_comparison}
\end{figure}

\paragraph{Packet Received and Fairness}
As shown in Fig. \ref{fig:overall_comparison}(a), the distributed multi-agent approach has proven to be more effective than other solutions in terms of the number of packets received. Compared to a single agent, the increase is $3.13\%$, $33.91\%$, $34.29\%$, and $53.74\%$, for VO, VI, HD, and BE respectively. This lies in the fact that in the distributed learning approach there is less data exchange between the AVs and the edge server. This results in a lower number of packet collisions and re-transmission. Furthermore, the fairness, in Fig. \ref{fig:overall_comparison}(b), in comparison between distributed learning and single agent is maintained for VO around $0.95$. For VI, it has a decrease of $14.7\%$, for HD Map an increase of $3.6\%$, and $1.37\%$ for BE. Regarding the multi-agent ACs, the distributed solution presented higher values in fairness. For VO there is an improvement of $0.8\%$, for VI $1.35\%$, for HD Map $21.7\%$, and $63\%$ for BE. This demonstrated that having each vehicle as an agent, especially in distributed learning the network performance is improved.

\section{Conclusion}
In this paper, we assessed the scalability of an application utilizing a Q-learning single-agent solution in a distributed multi-agent environment. More specifically, the sojourn time, used by a single-agent solution, can be scaled into vehicular ad hoc networks (VANET). Experimental results demonstrated that the multi-agent approach in terms of latency outperformed the single agent in four services by $40.4\%$, $36\%$, $43\%$, and $12\%$ for VO, VI, HD Map, and BE respectively. These superior performances imply the effectiveness of assigning the same reward function to each agent which provides the network performance feedback without any exchange of information between the agents. Taking the computational capacity into account, it is recommended to employ a centralized learning approach to mitigate the increase in computational costs in AVs, since the outcomes are promising when compared with those of a single agent. In scenarios where AVs have substantial computational capacity, distributed learning becomes feasible and is recommended due to the reduction in data exchange between the AVs and edge servers. 

In our future work, we will investigate the implementation of the proposed approach in an environment with multiple edge servers. Besides, a multi-agent heterogeneous environment where agents possess diverse objectives will also be further explored.

\bibliographystyle{IEEEtran}
\bibliography{IEEEabrv,Bibliography}

\begin{thebibliography}{10}
\providecommand{\url}[1]{#1}
\csname url@rmstyle\endcsname
\providecommand{\newblock}{\relax}
\providecommand{\bibinfo}[2]{#2}
\providecommand\BIBentrySTDinterwordspacing{\spaceskip=0pt\relax}
\providecommand\BIBentryALTinterwordstretchfactor{4}
\providecommand\BIBentryALTinterwordspacing{\spaceskip=\fontdimen2\font plus
\BIBentryALTinterwordstretchfactor\fontdimen3\font minus
  \fontdimen4\font\relax}
\providecommand\BIBforeignlanguage[2]{{%
\expandafter\ifx\csname l@#1\endcsname\relax
\typeout{** WARNING: IEEEtran.bst: No hyphenation pattern has been}%
\typeout{** loaded for the language `#1'. Using the pattern for}%
\typeout{** the default language instead.}%
\else
\language=\csname l@#1\endcsname
\fi
#2}}

\bibitem{hdmap_review_creation}
Z.~Bao, S.~Hossain, H.~Lang, and X.~Lin, ``A review of high-definition map
  creation methods for autonomous driving,'' \emph{Engineering Applications of
  Artificial Intelligence}, vol. 122, p. 106125, 2023.

\bibitem{nvidea_hdmap}
{ZVI GREENSTEIN (NVIDIA)}, ``{Announcing NVIDIA DRIVE Map: Scalable,
  Multi-Modal Mapping Engine Accelerates Deployment of Level 3 and Level 4
  Autonomy},'' online, accessed 8/15/23, March 2022,
  \url{"https://blogs.nvidia.com/blog/2022/03/22/drive-map-multi-modal-mapping-engine/"}.

\bibitem{sae_level5}
{SAE}, ``{SAE Levels of Driving Automation™ Refined for Clarity and
  International Audience},'' online, accessed 8/15/23, May 2021,
  https://www.sae.org/blog/sae-j3016-update.

\bibitem{chameleon}
C.~Zhu, Y.-H. Chiang, A.~Mehrabi, Y.~Xiao, A.~Ylä-Jääski, and Y.~Ji,
  ``Chameleon: Latency and resolution aware task offloading for visual-based
  assisted driving,'' \emph{IEEE Transactions on Vehicular Technology},
  vol.~68, no.~9, pp. 9038--9048, 2019.

\bibitem{edgeMap}
Q.~Liu, Y.~Zhang, and H.~Wang, ``Edgemap: Crowdsourcing high definition map in
  automotive edge computing,'' in \emph{ICC 2022 - IEEE International
  Conference on Communications}, 2022, pp. 4300--4305.

\bibitem{hdmap_processing_time}
\BIBentryALTinterwordspacing
J.~Lee, K.~Lee, A.~Yoo, and C.~Moon, ``Design and implementation of
  edge-fog-cloud system through hd map generation from lidar data of autonomous
  vehicles,'' \emph{Electronics}, vol.~9, no.~12, 2020. [Online]. Available:
  \url{https://www.mdpi.com/2079-9292/9/12/2084}
\BIBentrySTDinterwordspacing

\bibitem{performance_analysis_HDMAP}
J.~Redondo, Z.~Yuan, and N.~Aslam, ``Performance analysis of high-definition
  map distribution in vanet,'' in \emph{2023 International Wireless
  Communications and Mobile Computing (IWCMC)}, 2023, pp. 55--60.

\bibitem{ieee80211_cw}
Y.~Wang, A.~Ahmed, B.~Krishnamachari, and K.~Psounis, ``Ieee 802.11p
  performance evaluation and protocol enhancement,'' in \emph{2008 IEEE
  International Conference on Vehicular Electronics and Safety}, 2008, pp.
  317--322.

\bibitem{ieee80211_cw2}
S.~Eichler, ``Performance evaluation of the ieee 802.11p wave communication
  standard,'' in \emph{2007 IEEE 66th Vehicular Technology Conference}, 2007,
  pp. 2199--2203.

\bibitem{q_learning_fairness}
A.~Pressas, Z.~Sheng, F.~Ali, and D.~Tian, ``A q-learning approach with
  collective contention estimation for bandwidth-efficient and fair access
  control in ieee 802.11p vehicular networks,'' \emph{IEEE Transactions on
  Vehicular Technology}, vol.~68, no.~9, pp. 9136--9150, 2019.

\bibitem{q_learning_edca_policy_RL}
M.~Shinzaki, Y.~Koda, K.~Yamamoto, T.~Nishio, and M.~Morikura, ``Reducing
  transmission delay in edca using policy gradient reinforcement learning,'' in
  \emph{2020 IEEE 17th Annual Consumer Communications and Networking Conference
  (CCNC)}, 2020, pp. 1--6.

\bibitem{knowledge_driven}
Q.~Qi, J.~Wang, Z.~Ma, H.~Sun, Y.~Cao, L.~Zhang, and J.~Liao,
  ``Knowledge-driven service offloading decision for vehicular edge computing:
  A deep reinforcement learning approach,'' \emph{IEEE Transactions on
  Vehicular Technology}, vol.~68, no.~5, pp. 4192--4203, 2019.

\bibitem{deep_rl_resource_v2v}
H.~Ye, G.~Y. Li, and B.-H.~F. Juang, ``Deep reinforcement learning based
  resource allocation for v2v communications,'' \emph{IEEE Transactions on
  Vehicular Technology}, vol.~68, no.~4, pp. 3163--3173, 2019.

\bibitem{multi_agent_survey}
I.~Althamary, C.-W. Huang, and P.~Lin, ``A survey on multi-agent reinforcement
  learning methods for vehicular networks,'' in \emph{2019 15th International
  Wireless Communications \& Mobile Computing Conference (IWCMC)}, 2019, pp.
  1154--1159.

\bibitem{adaptive_cw}
A.~Kumar, G.~Verma, C.~Rao, A.~Swami, and S.~Segarra, ``Adaptive contention
  window design using deep q-learning,'' in \emph{ICASSP 2021 - 2021 IEEE
  International Conference on Acoustics, Speech and Signal Processing
  (ICASSP)}, 2021, pp. 4950--4954.

\bibitem{adaptive_cw_framework}
M.~Aguilar~Igartua, V.~Carrascal~Fr{\'\i}as, L.~J. de~la Cruz~Llopis, and
  E.~Sanvicente~Gargallo, ``Dynamic framework with adaptive contention window
  and multipath routing for video-streaming services over mobile ad hoc
  networks,'' \emph{Telecommunication Systems}, vol.~49, pp. 379--390, 2012.

\bibitem{adaptive_edca}
M.~A. Salem, I.~F. Tarrad, M.~I. Youssef, and S.~M.~A. El-kader, ``An adaptive
  edca selfishness-aware scheme for dense wlans in 5g networks,'' \emph{IEEE
  Access}, vol.~8, pp. 47\,034--47\,046, 2020.

\bibitem{cw_deep_RL_ieee802.11ax}
W.~Wydmański and S.~Szott, ``Contention window optimization in ieee 802.11ax
  networks with deep reinforcement learning,'' in \emph{2021 IEEE Wireless
  Communications and Networking Conference (WCNC)}, 2021, pp. 1--6.

\bibitem{our_sojourn_single_agent}
N.~A. Jeffrey~Redondo, Zhenhui~Yuan, ``Enhancement of high-definition map
  update service through coverage-aware and reinforcement learning,''
  \emph{arXiv preprint arXiv:}, 2024.

\bibitem{multi_agent_resource_management_5G_Wifi6}
F.~Zhou, L.~Feng, M.~Kadoch, P.~Yu, W.~Li, and Z.~Wang, ``Multiagent rl aided
  task offloading and resource management in wi-fi 6 and 5g coexisting
  industrial wireless environment,'' \emph{IEEE Transactions on Industrial
  Informatics}, vol.~18, no.~5, pp. 2923--2933, 2022.

\bibitem{scalable_reward}
J.~Leike, D.~Krueger, T.~Everitt, M.~Martic, V.~Maini, and S.~Legg, ``Scalable
  agent alignment via reward modeling: a research direction,'' \emph{arXiv
  preprint arXiv:1811.07871}, 2018.

\bibitem{multi_agent_coordination_book}
A.~K. Sadhu and A.~Konar, \emph{Multi-Agent Coordination: a reinforcement
  learning approach}.\hskip 1em plus 0.5em minus 0.4em\relax John Wiley \&
  Sons, 2020.

\bibitem{multi_agent_the_answer}
\BIBentryALTinterwordspacing
Y.~Shoham, R.~Powers, and T.~Grenager, ``If multi-agent learning is the answer,
  what is the question?'' \emph{Artificial Intelligence}, vol. 171, no.~7, pp.
  365--377, 2007, foundations of Multi-Agent Learning. [Online]. Available:
  \url{https://www.sciencedirect.com/science/article/pii/S0004370207000495}
\BIBentrySTDinterwordspacing

\bibitem{scalable_multi_agent}
G.~Qu, Y.~Lin, A.~Wierman, and N.~Li, ``Scalable multi-agent reinforcement
  learning for networked systems with average reward,'' \emph{Advances in
  Neural Information Processing Systems}, vol.~33, pp. 2074--2086, 2020.

\bibitem{avaq_edca_new_ac}
M.~M. Hasan and M.~Arifuzzaman, ``Avaq-edca: Additional video access queue
  based edca technique for dense ieee 802.11 ax networks,'' in \emph{2019
  International Conference on Computer, Communication, Chemical, Materials and
  Electronic Engineering (IC4ME2)}, 2019, pp. 1--4.

\bibitem{low_latency_new_ac}
H.~Park and C.~You, ``Latency impact for massive real-time applications on
  multi link operation,'' in \emph{2021 IEEE Region 10 Symposium (TENSYMP)},
  2021, pp. 1--5.

\bibitem{dynamic_queue}
H.~Zhang, W.~Tian, and J.~Liu, ``Dq-edca: Dynamic queue management based edca
  mechanism for vehicle communication,'' in \emph{2018 14th International
  Wireless Communications \& Mobile Computing Conference (IWCMC)}, 2018, pp.
  1131--1136.

\bibitem{queue_IoV}
A.~J. Gopinath and B.~Nithya, ``Enhancement of ieee802.11p-based channel access
  scheme for internet of vehicle (iov),'' in \emph{2019 IEEE International
  Conference on Advanced Networks and Telecommunications Systems (ANTS)}, 2019,
  pp. 1--6.

\bibitem{logical_EDCA}
L.~M. Rekik and M.~Bourenane, ``Logical edca: A novel edca mechanism for ieee
  802.11 based networks,'' in \emph{2020 Second International Conference on
  Embedded \& Distributed Systems (EDiS)}, 2020, pp. 99--104.

\bibitem{cw_cooperative_commu}
N.~Septa and S.~Wagh, ``Cooperative communication for safety message
  dissemination in unsaturated networks with varying contention windows,''
  \emph{International Journal of Communication Systems}, p. e5508, 2023.

\bibitem{RL_cw_simple}
K.~Sanada, H.~Hatano, and K.~Mori, ``Simple reinforcement learning based
  contention windows adjustment for ieee 802.11 networks,'' in \emph{2023 IEEE
  20th Consumer Communications \& Networking Conference (CCNC)}.\hskip 1em plus
  0.5em minus 0.4em\relax IEEE, 2023, pp. 692--693.

\bibitem{fairness_bidirectional_qlearning}
P.~Wang and S.~Wang, ``A fairness-enhanced intelligent mac scheme using
  q-learning-based bidirectional backoff for distributed vehicular
  communication networks,'' \emph{Tsinghua Science and Technology}, vol.~28,
  no.~2, pp. 258--268, 2023.

\bibitem{omnetppOMNeTDiscrete}
``Omnet++ discrete event simulator omnetpp.org,'' \url{https://omnetpp.org/},
  [Accessed 19-08-2023].

\bibitem{omnetppINETFramework}
``Inet framework - inet framework,'' \url{https://inet.omnetpp.org/}, [Accessed
  19-08-2023].

\bibitem{SUMO}
P.~A. Lopez, M.~Behrisch, L.~Bieker-Walz, J.~Erdmann, Y.-P. Flötteröd,
  R.~Hilbrich, L.~Lücken, J.~Rummel, P.~Wagner, and E.~Wiessner, ``Microscopic
  traffic simulation using sumo,'' in \emph{2018 21st International Conference
  on Intelligent Transportation Systems (ITSC)}, 2018, pp. 2575--2582.

\bibitem{veins}
C.~Sommer, R.~German, and F.~Dressler, ``Bidirectionally coupled network and
  road traffic simulation for improved ivc analysis,'' \emph{IEEE Transactions
  on Mobile Computing}, vol.~10, no.~1, pp. 3--15, 2011.

\bibitem{veinsgym}
M.~Schettler, D.~S. Buse, A.~Zubow, and F.~Dressler, ``{How to Train your ITS?
  Integrating Machine Learning with Vehicular Network Simulation},'' in
  \emph{12th IEEE Vehicular Networking Conference (VNC 2020)}.\hskip 1em plus
  0.5em minus 0.4em\relax Virtual Conference: IEEE, 12 2020.

\bibitem{acceleration_deceleration_petrol_car}
\BIBentryALTinterwordspacing
P.~Bokare and A.~Maurya, ``Acceleration-deceleration behaviour of various
  vehicle types,'' \emph{Transportation Research Procedia}, vol.~25, pp.
  4733--4749, 2017, world Conference on Transport Research - WCTR 2016
  Shanghai. 10-15 July 2016. [Online]. Available:
  \url{https://www.sciencedirect.com/science/article/pii/S2352146517307937}
\BIBentrySTDinterwordspacing

\bibitem{dataset}
``Power bi report,'' \url{https://www.tinyurl.com/tadudashboard}, [Accessed
  22-08-2023].

\bibitem{dataRate_cisco_voice_szigeti2005end}
\BIBentryALTinterwordspacing
T.~Szigeti and C.~Hattingh, \emph{End-to-end Qos Network Design}, ser. Cisco
  Press networking technology series.\hskip 1em plus 0.5em minus 0.4em\relax
  Cisco Press, 2005. [Online]. Available:
  \url{https://books.google.co.uk/books?id=WOPoD6cGXEsC}
\BIBentrySTDinterwordspacing

\bibitem{cisco_2}
{Cisco}, ``{QoSVoIP},''
  {https://www.cisco.com/c/en/us/td/docs/ios/solut-ions\_docs/qos\_solutions/QoSVoIP/QoSVoIP.pdf},
  November 2023, online, Accessed 14/11/2023.

\bibitem{dataRateVideo_googleYouTube}
{YouTube. ''YouTube recommended upload encoding settings - YouTube Help.''
  support.google.com.}
  {https://support.google.com/youtube/answer/1722171?hl=en-\#zippy=\%2Ccontainer-mp\%2Caudio-codec-aac-lc\%2Cvideo-codec-h\%2Cframe-rate\%2Cbitrate}.
  (Accessed Aug. 22, 2023).

\bibitem{5gaa_delay_2}
{5GAA Automotive Association}, ``{C-V2X Use Cases Volume II: Examples and
  Service Level},'' online, Accessed 15/08/23, October 2020,
  \url{"https://5gaa.org/content/uploads/2020/10/5GAAWhite-PaperC-V2X-Use-Cases-Volume-II.pdf"}.

\bibitem{fairness_jain}
R.~K. Jain, D.-M.~W. Chiu, W.~R. Hawe, \emph{et~al.}, ``A quantitative measure
  of fairness and discrimination,'' \emph{Eastern Research Laboratory, Digital
  Equipment Corporation, Hudson, MA}, vol.~21, 1984.

\end{thebibliography}

\vspace{12pt}

\end{document}